\documentclass[]{interact}

\usepackage{epstopdf}
\usepackage[caption=false]{subfig}

\usepackage[numbers,sort&compress]{natbib}
\bibpunct[, ]{[}{]}{,}{n}{,}{,}
\renewcommand\bibfont{\fontsize{10}{12}\selectfont}
\makeatletter
\def\NAT@def@citea{\def\@citea{\NAT@separator}}
\makeatother

\theoremstyle{plain} 

\theoremstyle{definition}

\theoremstyle{remark}

\usepackage{listings}
\usepackage{graphicx}
\usepackage{multirow}
\usepackage{booktabs}
\usepackage{array}
\usepackage{hyperref}

\usepackage{lmodern}  
\usepackage{amsmath}  
\usepackage{xcolor}   

\makeatletter
\newcommand{\footnoteref}[1]{%
\ltx@ifpackageloaded{hyperref}{
  \ifHy@hyperfootnotes
    \hbox{\hyperref[#1]{%
            \@textsuperscript {\normalfont \ref*{#1}}}}%
  \else
    \hbox{\@textsuperscript {\normalfont \ref*{#1}}}%
  \fi%
}{
    \hbox{\@textsuperscript {\normalfont \ref{#1}}}%
 }%
}
\makeatother

\begin{document}

\articletype{RESEARCH PAPER}

\title{Mining experimental data from Materials Science literature with Large Language Models: an evaluation study}

\author{
    \name{Luca Foppiano\textsuperscript{a,b}\thanks{Corresponding authors: Luca Foppiano (luca@foppiano.org), Guillaume Lambard (LAMBARD.Guillaume@nims.go.jp), and Masashi Ishii (ISHII.Masashi@nims.go.jp)}, Guillaume Lambard\textsuperscript{c}, Toshiyuki Amagasa\textsuperscript{b}, Masashi Ishii\textsuperscript{c}}
    \affil{\textsuperscript{a}Materials Modeling Group, Center for Basic Research on Materials, National Institute for Materials Science, Ibaraki-ken, Tsukuba-shi, 1-1 Namiki, 305-0044, Japan \\
    \textsuperscript{b}Knowledge and Data Engineering, Centre for Computational Sciences, University of Tsukuba, JP \\
    \textsuperscript{c}Data-driven Materials Design Group, Center for Basic Research on Materials, National Institute for Materials Science, Ibaraki-ken, Tsukuba-shi, 1-1 Namiki, 305-0044, Japan 
    }
}

\maketitle

\begin{abstract}
This study is dedicated to assessing the capabilities of large language models (LLMs) such as GPT-3.5-Turbo, GPT-4, and GPT-4-Turbo in extracting structured information from scientific documents in materials science. 
To this end, we primarily focus on two critical tasks of information extraction: (i) a named entity recognition (NER) of studied materials and physical properties and (ii) a relation extraction (RE) between these entities. 
Due to the evident lack of datasets within Materials Informatics (MI), we evaluated using SuperMat, based on superconductor research, and MeasEval, a generic measurement evaluation corpus. 
The performance of LLMs in executing these tasks is benchmarked against traditional models based on the BERT architecture and rule-based approaches (baseline).
We introduce a novel methodology for the comparative analysis of intricate material expressions, emphasising the standardisation of chemical formulas to tackle the complexities inherent in materials science information assessment. 
For NER, LLMs fail to outperform the baseline with zero-shot prompting and exhibit only limited improvement with few-shot prompting. 
However, a GPT-3.5-Turbo fine-tuned with the appropriate strategy for RE outperforms all models, including the baseline. Without any fine-tuning, GPT-4 and GPT-4-Turbo display remarkable reasoning and relationship extraction capabilities after being provided with merely a couple of examples, surpassing the baseline. 
Overall, the results suggest that although LLMs demonstrate relevant reasoning skills in connecting concepts, specialised models are currently a better choice for tasks requiring extracting complex domain-specific entities like materials. These insights provide initial guidance applicable to other materials science sub-domains in future work.
\end{abstract}

\section{Introduction}

Mining experimental data from literature has become increasingly popular in materials science due to the vast amount of information available and the need to accelerate materials discovery using data-driven techniques. Data for machine learning in materials science is often sourced from published papers, material databases, laboratory experiments, or first-principles calculations~\cite{xuSmallDataMachine2023}. The introduction of big data in materials research has shifted from traditional random techniques to more efficient, data-driven methods. Data mining of computational screening libraries has been shown to identify different classes of strong $\rm CO_{2}$-binding sites, enabling materials to exhibit specific properties even in wet flue gases~\cite{boyd2019data}. Machine learning techniques have been employed for high-entropy alloy discovery, focusing on probabilistic models and artificial neural networks~\cite{rao2022machine}. However, the use of advanced machine learning algorithms in experimental materials science is limited by the lack of sufficiently large and diverse datasets amenable to data mining~\cite{zakutayev2018open}. A present central tenet of data-driven materials discovery is that with a sufficiently large volume of accumulated data and suitable data-driven techniques, designing a new material could be more efficient and rational~\cite{huan2016polymer}. The materials science field is moving away from traditional manual, serial, and human-intensive work towards automated, parallel, and iterative processes driven by artificial intelligence, simulation, and experimental automation~\cite{pyzer2022accelerating, huber2020machine}. But, materials science literature is a vast source of knowledge that remains relatively unexplored with data mining techniques~\cite{park2021advances}, especially for the reason that materials science data come in diverse forms such as unstructured textual content and structured tables and graphs, adding complexity to the extraction process. As a result, nowadays, many projects still depend on manual data extraction. While extensive structured databases contain accumulated experimental data~\cite{chittambigdat2021}, they remain limited in number and highly costly due to the amount of human labour involved~\cite{madataug2020}.

Additionally, addressing issues related to the quality and meaning of materials science data often demands a curation step assisted by a sub-domain knowledge frequently specific to the approached sub-field of materials science, e.g. polymers, metal-organic frameworks, high-entropy alloys, etc., with their own physical and chemical phenomena, methods and protocols, terminology and jargon. For instance, the classification of superconductors can be complex and sometimes arbitrary, blending compound-based classes like cuprates~\cite{parinov2013microstructure} and iron-based~\cite{hosono2015exploration} materials with unconventional classes like heavy fermions~\cite{mydeen2020electron}. The classification of superconductors can also be based on phenomena such as the Meissner effect, which describes how superconductors expel magnetic fields~\cite{bardeen1957theory}. Superconductors can be divided into two classes according to how this breakdown occurs, the so-called type-I and type-II superconductors. As these classifications are not mutually exclusive certain materials could potentially fall into multiple categories, for example, a material can be both a cuprate and a type-II superconductor, the classification of superconductors is a complex task that demands an extensive knowledge of the developments and current state-of-the-arts.
Moreover, substantial confusion may occur due to the cross-domain polysemy of used words, terms and symbols. In different sub-domains, the same term can take on specific nuances or meanings unique to a given sub-domain. This phenomenon is common in language and can lead to misunderstandings if the context of the sub-domain is unclear. For instance, the acronym "TC" or "T\textsubscript{c}" will be employed for denoting a "Temperature Curie" or a "superconducting critical temperature", respectively. These sub-domain-specific conventions pose a significant challenge when attempting to create structured datasets across various sub-domains effectively.

Meanwhile, the advent of large language models (LLMs) has inaugurated a new technological era marked by extraordinary potential. These models not only excel in linking diverse concepts but also in engaging in sophisticated conversational reasoning~\cite{zhang2023one,yao2023tree,valmeekam2023planning,sun2023pearl}. 
In comparison, rule-based approaches are simpler and faster (tokens per second); however, they are time-intensive to fine-tune and have weak generalisation capabilities, as new rules should be clarified on a case-by-case basis. 
Small Language Models (SLMs), e.g. BERT-based models, are more specific to the task on which they are pre-trained. The data size used for pre-training LLMs is usually high enough to contain a high diversity of contexts, thus necessitating fewer examples at the fine-tuning stage than SLMs models.  

LLMs offer the possibility of integrating large corpus of textual data at training, with often the ability to ingest large textual inputs at inference with a context window ranging from 4,096 to 128,000 tokens for GPT-3.5-Turbo and GPT-4-Turbo~\cite{OpenAIDocJan2024}, respectively (at the time of this writing). 
Differently, BERT-based encoders are limited to only 512 tokens, whereas 1,000 tokens are about 750 English words. 
The size of BERT models does not allow them to sustain their contextual memory after fine-tuning. LLMs possess the capacity to be fine-tuned and retain the contextual knowledge from pre-training, which gives them an advantage in terms of generalisation to other datasets.
Finally, the interaction with LLMs via prompts, i.e., tailored instructions, changes the construction paradigm of programmatic solutions, making them more accessible, flexible, and suitable to human operators. Nevertheless, the actual capabilities of LLMs in reasoning, understanding, and recognition are still constantly evolving and being evaluated. 

Previous studies in Information Extraction (IE) have shown evidence of LLMs proficiency in general tasks, presenting a valuable opportunity to develop more flexible Text and Data Mining (TDM) processes. Still, they fall short in areas where specific knowledge is required~\cite{kokon2023chatgpt}. In particular, LLMs are on par with SLMs in most of the discriminative tasks such as named entity recognition (NER), relation extraction (RE) and event detection (ED) in general domains~\cite{ma2023large}, in history~\cite{gonzalez2023yes}, and biology~\cite{moradi2021gpt}. 
Other works testing chemistry capabilities found that GPT-4 understands various aspects of chemistry, including chemical compounds~\cite{hatakeyama2023prompt}; however, its knowledge is general and lacks methods for learning through retrieving recent literature~\cite{hatakeyama2023using}.

Therefore, this study assesses LLMs' ability to comprehend, manipulate, and reason with complex information that demands substantial background knowledge, as in materials science. 

The objectives of this work can be summarised with the two following questions: 
\begin{itemize}
    \item \textbf{Q1}: \textit{How effectively can LLMs extract materials science-related information?}
    \item \textbf{Q2}: \textit{To what extent can LLMs use reasoning to relate complex concepts?}
\end{itemize}

\vspace{.1cm}
We first classify the fundamental components of the materials science knowledge directed towards designing novel materials with functional properties into two main entity classes: material and property expressions. 
Properties, e.g., a critical temperature of 4 K, are expressed using measurements of physical quantities. 
They exhibit a structured format, including modifiers (e.g., "between", "less than", "approximately", or symbols such as "$>$" or "$\sim$"), values, and units, with a wide range of potential values. 
In contrast, material definitions are conceptually loose and often depend on the specific domain. They may require a substantial amount of accompanying text for a comprehensive description, encompassing details, e.g., compositional ratios, doping agent and ratio, synthesis protocol, process, and additional adjunct information. 
From a fundamental compositional standpoint, materials are defined by their chemical formula. However, in practice, authors in literature may frequently employ substantives such as commercial names, well-known terms, or crafted designations to describe samples, all of which streamline information in their research papers. Nonetheless, conveying such definitions can unambiguously be challenging.

To address Q1, we evaluate the LLM's performance on NER tasks related to materials and properties extraction. For each task, we choose a pertinent dataset and analyse the performance of each LLM.

Named Entity Recognition (NER)~\cite{nadeau2007survey}, alternatively referred to as named entity identification or entity extraction, stands as a pivotal component within information extraction. Its primary objective is to pinpoint and categorise named entities within unstructured text, assigning them to predefined categories such as individual names, organisations, geographic locations, medical codes, temporal expressions, quantities, monetary values, percentages, and more. The process of identifying entities aligns closely with sequence labelling tasks, wherein a string of text undergoes analysis, and each token within it (basic unit of text processing, typically a word or a sequence of characters that is treated as a single unit) is designated to one of the pre-established categories. For instance, these categories may include material, doping, condition, or property, among others.


We address Q2 by assessing the capability to establish connections between a predefined set of entities and extract relationships within a given context. 
Extracting relations between entities is a foundational undertaking in NLP. It entails discerning connections or associations among entities referenced within textual data. 
For instance, in biomedical research, relationship extraction might involve identifying the association between specific genes and diseases mentioned in scientific literature.

In both cases, we compare the outcomes against a baseline determined by scores (Precision, Recall and F1-score) achieved on the same datasets by either a BERT-based encoder or a rule-based algorithm we have developed in a previous work~\cite{lfoppiano2023automatic, foppiano2019quantities}.
Our requirement for the models to be capable of generating output in a valid JSON (JavaScript Object Notation) format is part of our efforts to extract structured databases (Section~\ref{subsubsec:output-format}).
 

The evaluation of generative models brings an additional complexity. 
Traditional SLM implementations for solving NER tasks are based on sequence labelling algorithms. They classify each token in the input stream with a limited number of labels, returning a sequence that fits the original input (same number of tokens and structure). 
Evaluating their performance against expected datasets involves a straightforward comparison of values. Soft-matching techniques can be employed to overlook minor discrepancies. 
However, with generative models, the output tokens may be structured in ways that significantly differ from the original input sequence. 
In more general scenarios, semantic models that compare the vectorised representations of two sequences can be utilised~\cite{reimers2019sentencebert}. 
Nevertheless, when dealing with concepts like material expressions, a specialised approach is needed. 
As an illustration, the terms "solar cell" and "solar cells" represent identical concepts. Yet, the materials denoted by "Ca" (Calcium) and "Cr" (Chromium) are entirely distinct, highlighting a difference of just one letter between the two examples.
For this reason, we introduce a novel evaluation method for material names, which involves normalising materials to their chemical formulas before conducting a pairwise comparison of each element. This approach provides a more meaningful and context-aware assessment of the model's performance.

We summarise our contributions as follows: 

\begin{itemize}
    \item We designed and ran a benchmark for LLMs on information extraction, particularly NER of materials and properties. This contribution addresses Q1. 
    \item We evaluated LLMs on RE on entities in the context of materials science to address Q2.
    \item We propose a novel approach for evaluating Information Extraction tasks applied to materials entities which leverage "formula matching" via pairwise element comparison.
\end{itemize}

\section{Method}
\label{sec:method}

We chose three OpenAI LLM models reported with their specific names for performing API calls: GPT-3.5-Turbo (gpt-3.5-turbo-0611), GPT-4 (gpt-4), and GPT-4-Turbo (gpt-4-0611-preview). 
The consideration of open-source LLMs has been deferred to future work due to their limited capability to generate output in a valid JSON format (Section~\ref{subsubsec:output-format}, necessitating a more in-depth investigation.

Our evaluation uses different strategies: zero-shot prompting, few-shot prompting, and fine-tuning (or instruction-learning).  
Few-shot prompting refers to the model's ability to adapt and perform a new task with minimal examples or prompts. In contrast, zero-shot prompting denotes the model's capability to generalise to tasks it has not been explicitly trained on, emphasising transfer learning within the language domain.
Finally, fine-tuning involves adjusting the parameters of a pre-trained model on a specific task or domain using a smaller, task-specific dataset to enhance its performance for that particular application.

We selected two datasets for evaluation: MeasEval~\cite{harper2021semeval2021}, a SemEval 2021 task of extracting counts, measurements, and related context from scientific documents and SuperMat, an annotated and linked dataset of research papers on superconductors~\cite{lfoppiano2021supermat}. 
SuperMat contains both materials and properties and, for copyright reasons, is not publicly distributed. 
This reduces the risk that its annotations had been used during the pre-training of any of the LLMs.

Baseline scores were established using a SciBERT-based~\cite{beltagy2020scibert} encoder and RE rule-based algorithm~\cite{lfoppiano2023automatic} for material-related extractions. Grobid-quantities~\cite{foppiano2019quantities} served as the baseline for NER on properties extraction evaluated against MeasEval.

Evaluation scores, encompassing Precision, $\rm TP/(TP+FP)$, Recall, $\rm TP/(TP+FN)$), and F1-score, $\rm 2\, Precision \times Recall / (Precision + Recall)$, were derived from pairwise comparisons between predicted and expected entities, where TP, FP and FN are the true positive, false positive and false negative instances, respectively. Precision gauges accuracy, recall assesses information capture, and F1-Score is their harmonic mean.

The evaluations condense average F1 scores and their standard deviation over three extraction runs. The raw tables with all detailed scores are provided in Appendix~\ref{appendix:full-evaluation-results}.

\subsection{Named Entities Recognition}
\label{sec:ner}
The NER task consists of identifying relevant entities: materials, expressed through a multitude of expressions~\cite{lfoppiano2021supermat}, or properties, expressed as measurements of physical quantities~\cite{foppiano2019quantities}. 

We calculated the evaluation scores using four different matching approaches. However, we will present only the most relevant to the task (leaving the complete tables\footnote{The calculation of micro average provides a measure independent of the distribution of the extracted entities over the different documents.} in Appendix~\ref{appendix:full-evaluation-results}): 
\begin{itemize}
    \item \textbf{strict}: Exact matching
    \item \textbf{soft} Matching using Ratcliff/Obershelp~\cite{ratcliff_obershelp} with a threshold at 0.9\footnote{\label{ref:threshold}The threshold is fixed to a value yielding more than 90\% similarity.}
    \item \textbf{Sentence BERT} Comparison using semantic similarity of sequences using Sentence BERT with a cross-encoder~\cite{reimers2019sentencebert}, applying a threshold set at 0.9\footnoteref{ref:threshold}
    \item \textbf{formula matching} Our novel method compares material expressions via formula normalisation and element-by-element exact matching.
\end{itemize}

Prompts for interacting with LLMs are defined by two components: system and user prompts. 
The system prompt is the initial instruction guiding the model's output generation, defining the task or information sought. In contrast, the user prompt is the user's input, specifying their request and shaping the model's response.

The system prompt below was fixed across all tasks. It was specifically crafted to prevent the creation of non-existing facts and favour standardised answers (e.g., "I don't know," "None," etc.) in case of inability to respond.

\begin{lstlisting}[caption=Generic system prompt common to all requests]
Use the following pieces of context to answer the user's question. 
If you don't know the answer, just say that you don't know, don't try to make up an answer. 
----------------
{text}
\end{lstlisting}

The users' prompts for NER with zero-shot prompting were described including the definitions and examples from the SuperMat\footnote{\url{https://supermat.readthedocs.io}} and MeasEval\footnote{\url{https://github.com/harperco/MeasEval/tree/main/annotationGuidelines\#basic-annotation-set}} annotations guidelines, respectively.  

Below are the user prompt templates used for both materials and properties extraction: 

\begin{lstlisting}[caption=User prompt designed for extracting materials and properties. The entity descriptions are separated by dashed lines ("\texttt{------}").]
What are the superconductor materials mentioned in the text? 
Only provide the mention of the materials. Avoid repetition. 

The material can be expressed as follows:
- chemical formula with variables not substituted, like La(1-x)Fe(x),
- chemical formula with substitution variables like Zr 5 X 3 (X = Sb, Pb, Sn, Ge, Si and Al)
- with complete or partial abbreviations like (TMTSF) 2 PF 6,
- doping rates are represented as variables (x, y or other letters) appearing in the material names. These values can be used to complement the material variables (e.g. LaFexO1-x).
- doping rates as percentages, like 4% Hdoped sample or 14% Cu doped sample
- material chemical form with no variables e.g. LaFe03NaCl2 where the doping rates are included in the name
- chemical substitution or replacements, like (A is a random variable, can be any symbol): A = Ni, Cu, A = Ni, Ni substituted (which means A = Ni)
- chemical substitution with doping ratio, like (A is a random variable, can be any symbol): A = Ni and x = 0.2

If you don't know the answer, just say you don't know, don't try to make up an answer.

-----

Quantity is either a Count, consisting of a value, or a Measurement, 
consisting of a value and usually a unit. A Quantity can additionally include optional Modifiers like tolerances.
Include relevant text that indicates the application of a modifier, such as "between", "less than", "approximately", 
or symbols such as ">" or "~" if they are contiguous with the span. Ignore them if they are separated by additional text.
 
Example: "The soda can's volume was 355 ml", the quantity is "355 ml".

Extract all the Quantities in the text.
\end{lstlisting}

Then, we applied a few-shot prompting technique by incorporating in the users' prompt template above a set of suggestions extracted from the text (see Listing 3 below) using the respective SLMs based on the fine-tuned SciBERT-encoder for materials and properties, i.e., grobid-superconductors~\cite{lfoppiano2023automatic} and grobid-quantities~\cite{foppiano2019quantities}, respectively. Also, as these suggestions originate from another model, they may not be entirely accurate; hence, we emphasised in the prompts that they only serve as examples or hints that the LLMs may ignore.  

\begin{lstlisting}[caption=Few-shot prompting modified prompt template.]
[...]
Here are some examples appearing in the text: {hints}
[...]
\end{lstlisting}

\subsubsection{Output format}
\label{subsubsec:output-format}

For all tasks, we required the output to be formatted using a valid JSON document. 
We justify this decision for three main reasons: 
a) The responses need to be machine-readable so that the de-serialisation from JSON to objects in many programming languages becomes a trivial operation (e.g., Python, JavaScript).
b) The JSON schema can be defined through a documented format regardless of the programming language or platform. 
Finally, c) the JSON format is an open standard that can be used by anyone and does not require reinventing the wheel by re-implementing any transformation steps from scratch.

The JSON output was obtained by adding formatting instructions in the user's prompt based on the expected output data model, for which different concepts were described differently (e.g., properties are described as a value and an optional unit).
We used the implementation provided by the LangChain library\footnote{\url{https://github.com/langchain-ai/langchain}} of which one example is illustrated as follows. 

\begin{lstlisting}[caption=Example of formatting instruction to a valid JSON format]
The output should be formatted as a JSON instance that conforms to the JSON schema below.

As an example, for the schema {"properties": {"foo": {"title": "Foo", "description": "a list of strings", "type": "array", "items": {"type": "string"}}}, "required": ["foo"]}
the object {"foo": ["bar", "baz"]} is a well-formatted instance of the schema. The object {"properties": {"foo": ["bar", "baz"]}} is not well-formatted.

Here is the output schema:
```
{"properties": {"material": {"title": "Material", "description": "Material or sample name, chemical formula, acronym. Include everything that describes the material.", "type": "string"}, "material_extra_info": {"title": "Material Extra Info", "description": "Additional information about the material", "type": "string"}}, "required": ["material"]}
```
\end{lstlisting}

\subsubsection{Formula matching}

Matching materials poses challenges with generative models, while encoder and sequence labelling models maintain unchanged the output from the input sequences. Therefore, evaluating generative models can be complex due to potentially divergent yet semantically equivalent output sequences. 
Previous works~\cite{taylor2022galactica} resort to manual evaluation due to these challenges. Notably, as of the time of writing, no specialised approach tailored for material expressions existed. 
Utilising Sentence BERT, trained on general text, does not ensure accurate material embeddings, raising concerns about the meaningfulness of final matches. 
To address issues arising from variable sets and to enhance evaluation precision, we propose a novel method named \textit{formula\_matching}, involving element-by-element pairwise comparisons on normalised formulas for extracted material denominations.



\begin{figure}[ht]
  \centering
  \includegraphics[width=1\textwidth]{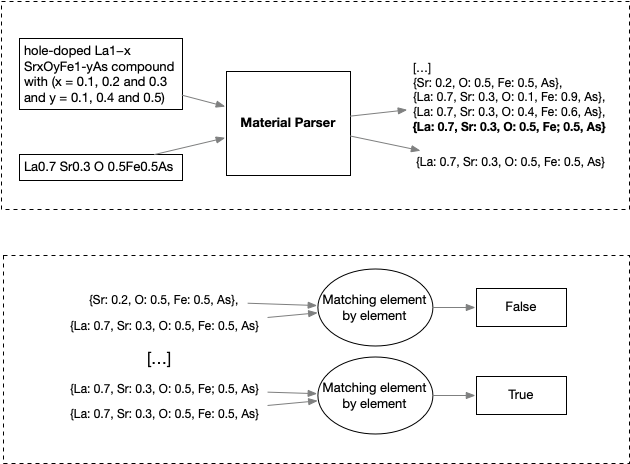} 
  \caption{Two materials that appear to have a very different composition are, in reality, overlapping. (Top) Summary of the Material Parser. More information is available in~\cite{lfoppiano2023automatic}. (Bottom) The pairwise comparison of each chemical formula is performed element-by-element.  }
  \label{fig:formula-matching-schema}
\end{figure}

This approach extends strict matching and is activated only when the two input strings differ. In such instances, as depicted in Figure~\ref{fig:formula-matching-schema}, the material expressions slated for comparison undergo normalisation to their formulas using a material parser developed in our prior work~\cite{lfoppiano2023automatic} (Figure~\ref{fig:formula-matching-schema} top). 
The material parser is adept at handling noisy material expressions and strives to parse them effectively. The anticipated output includes a structured representation with the chemical formula presented as a raw string and a dictionary detailing elements and their respective amounts. 
Subsequently, these structures are compared element by element, as depicted in Figure~\ref{fig:formula-matching-schema} bottom.
The summarised evaluation scores described in Section~\ref{sec:results-ner-materials} are calculated using the formula matching. 

Evaluation and discussion of this method are detailed in Section~\ref{subsec:formula-matching}.

\subsection{Relation Extraction}
\label{sec:re}
The baseline is established by a rule-based algorithm from our previous work~\cite{lfoppiano2023automatic}, which was evaluated with SuperMat and for which we report the aggregated result in Section~\ref{sec:re}. 

The prompts are designed by providing a list of entities and requesting the LLM to group them based on their relation. 
Unlike the NER task, the LLM is expected to reuse information passed in the prompt to compose the response: non-matching information is considered incorrect.
The summarised scores in Section~\ref{sec:results-re} are obtained with strict matching. 

The previous considerations remain relevant for both system and user prompts, with the task description reiterated in each prompt.

\begin{lstlisting}[caption=System prompt for RE modified by emphasising the tasks]
You are a useful assistant, who knows about materials science, physics, chemistry and engineering.
You will be asked to compute relation extraction given a text and lists of entities. 
If you are not sure, don't try to make up your answer, just answer "None". 
\end{lstlisting}

We add specific rules to avoid creating invalid groups of relations and to ignore responses containing entities not supplied in the user prompt or empty relation blocks. 


    

The prompt for few-shot prompting was assembled by injecting three examples listed between the dashed lines ("\texttt{--------}") in the zero-shot prompt:

\begin{lstlisting}[caption=Few-shot prompting for extracting relations from lists of entities]
Given a text between triple quotes and a list of entities, find the relations between entities of different classes: 
"""
{text}
"""

{entities}
 
Use the following examples separated by "--------" to learn the task: 
--------
text 1: The researchers of Mg have discovered that MgB2 and MgB3 are superconducting at 29-31 K at ambient pressure.

entities 1:
 materials: MgB2, Mg, MgB3
 tcs: 29-31 K
 pressure: ambient pressure
 
Result 1: 
 material: MgB2, 
 tc: 29-31K, 
 pressure: ambient pressure:
 
 material: MgB3, 
 tc: 29-31K, 
 pressure: ambient pressure:

--------
Text 2: We are studying the material La 3 A 2 Ge 2 (A = Ir, Rh). The critical temperature T C = 4.7 K discovered for La 3 Ir 2 Ge 2 in this work is by about 1.2 K higher than that found for La 3 Rh 2 Ge 2.

entities 2:
 materials: La 3 A 2 Ge 2 (A = Ir, Rh), La 3 Ir 2 Ge 2, La 3 Rh 2 Ge 2
 tcs: 4.7 K, 1.2 K
 
Result 2: 
 material: La 3 Ir 2 Ge 2
 tc: 4.7 K

--------
Text 3: The experimental discovery of the high-temperature superconducting state in the compressed hydrogen and sulfur systems H2S (TC = 150 K for p = 150 GPa) and H3S (TC = 203 K for p = 150 GPa)

entities 3:
 materials: H2S, H3S
 tcs: 150 K, 203 K
 pressures: 150 GPa, 150 GPa
 
Result 3: 
 material: H2S,
 tc: 4.7 K,
 pressure: 150 GPa
 
 material: H3S,
 tc: 150 K,
 pressure: 150 GPa
--------

Apply strictly the following rules:  
    - if material is not specified, ignore the relation block,
    - if tc is not specified in absolute values, ignore the relation block 
\end{lstlisting}


\subsubsection{Shuffled vs non-shuffled evaluation}
\label{subsub:shuffled-non_shuffled-eval}
The list of entities supplied to the Language Model (LLM) might be derived based on their order of appearance, creating a scenario where a model generating relations sequentially may achieve an inflated score that does not accurately reflect its relational inference capabilities. 
To address this, we evaluate each model for RE using two strategies: a \emph{non-shuffled evaluation}, where entities are presented in the order they appear in the original document, and a \emph{shuffled evaluation}, where entities are randomly rearranged before being introduced to the prompt.

\subsection{Consideration about the fine-tuning}
\label{subsec:consideration-fine-tuning}

We fine-tuned the GPT-3.5-Turbo model using the OpenAI platform, which ingested training and testing data and generated a new model in a few hours. 
At the time of writing this article, the fine-tuning of GPT-4 and GPT-4-Turbo is not available. 
All fine-tuned models were trained using the default parameters selected by the OpenAI platform.

Table~\ref{tab:amount-data-fine-tuned} illustrates the dimension of each dataset. 
The fine-tuned model for properties extraction was trained using the "grobid-quantities dataset"~\cite{foppiano2019quantities} because MeasEval did not contain enough examples for a consistent and unbiased evaluation. 


The primary challenge encountered when employing a fine-tuned model was to achieve a valid, machine-readable JSON format. 
Therefore, we formatted the training data with an expected output in valid JSON format. 
However, the obtained fine-tuned model struggled to produce valid JSON in its output, leading us to hypothesise that this limitation might be attributed to a shortage of training examples. 
To address this, we modified our training data expected output from JSON to a pseudo format structured with spaces and break-lines, facilitating simpler handling by the model. The subsequent example illustrates the expected output for a RE task:

\begin{lstlisting}[caption=Example format of the expected answer for the RE task]
    material: mat1, tc: 22K, 
    material: mat2, tc: 24K, pressure: 2GPa
\end{lstlisting}

We followed the same approach for fine-tuning the model for the NER task: 

\begin{lstlisting}[caption=Example format of the expected answer for the NER task]
    materials: 
     - material1
     - material2
     - material3
\end{lstlisting}

Using this technique, we could fine-tune a model and shape its behaviour to answer conversationally. Then, we used the GPT-3.5-Turbo base model to transform the response into JSON format. 

To fine-tune the model for the RE task, we introduced the sorting variability in the entity lists provided in the prompt (Section~\ref{sec:re}). This approach does not modify the size of the data set and reduces the possibility that the model learns to aggregate entities in the order they appear in the document. This is the default approach we define as "FT.base" compared to others.
In Section~\ref{subsubsec:data-variability}, we discuss the impact of two additional strategies for preparing the fine-tuning data. 
First, "FT.document\_order" keeps the lists of entities as they appear in the document. 
For example, the made-up sentence \textit{"The two materials MgB2 and MgB3 showed Tc of 39K and 40K, respectively"} will lead to two lists of entities "MgB2, MgB3" and "39K and 40K" which could be assigned in order (MgB2, 39K) and (MgB3, 40K).
Intuitively, this leads to poor performance, as we see when evaluating with shuffling conditions (Section~\ref{sec:results-re}). 
The second strategy, "FT.augmented", is to augment the size of the dataset, generating multiple training records with a further shuffled entity list for each example in "FT.base". The data used with this strategy is roughly double that of "FT.base" (Table~\ref{tab:amount-data-fine-tuned}). We expect this strategy to obtain similar or better results than "FT.base". 

\section{Results and discussions}
\label{sec:results}
In this section, we present and discuss the formula matching and the aggregated results of our evaluations for the LLMs. The completed raw results are available in the Appendix \ref{appendix}. 

\subsection{Limitation of this study}
In this paper, we aim to estimate how well LLMs work in tasks related to materials science. 
Due to the lack of clean datasets covering the entire materials science domain, we used a dataset that focuses on superconductor material. 
While our goal is to propose a methodology, we are aware that our results need to be verified empirically in other materials science sub-domains in future works. 
The following intuitions support our hypothesis: for material NER, we expect that the forms on which materials are presented in other domains would have similar expressions to the ones used in superconductor research, considering that chemical formulas, sample names, and commercial names would unlikely be very different between domains. 
Furthermore, the properties, expressed as measurement and physical quantities, are common to all domains; although the statistical distribution could be different, we don't expect dramatic differences within materials science. 
On the other hand, RE tasks surely require more datasets that focus both on different domains and different flavours of the same task. 
As an example, the MatSCIRe~\cite{mullick2024matscire} dataset, which covers battery-related research, proposes a structure that challenges the relation extraction only between two entities (binary extraction) with the addition of the type of relation which could be inferred by the properties being extracted. 
In conclusion, we will remand the generalisation for further work. 

\subsection{Formula matching}
\label{subsec:formula-matching}
We evaluated the formula matching to measure two main pieces of information: the gain in the F1-score, and the correctness, as the number of invalid new matches, of the gain. 
We compared the formula matching with the strict matching because a) it is simple to reproduce and understand visually, and b) the formula matching is built on top of strict matching. 
We would have more difficulties explaining matches provided by soft matching or SentenceBERT. 

We examined the GPT-3.5-Turbo NER extraction (discussed in Section~\ref{sec:results-re}). 107 out of the 1402 expected records matched correctly using strict matching (P: 22.5\%, R: 13.64\%, F1: 17.01\%). Applying formula matching on the mismatching records, we obtained an additional 176 matches (P: 61.12\%, R: 36.00\%, F1: 45.31\%), for a total gain in F1-score of 28.3 (+266\%). 
For the new 176 records that the formula matching was identifying, we manually examined each pair finding 5 incorrect matches, which corresponds to an error rate of 2.5\%. 

Most of the mismatches in the strict matching caught up by the formula matching were due to missing adjoined information. 
The LLMs were not able to include information about doping or shape in the response (e.g. \texttt{hole-doped La 2-x Sr x CuO 4} was not matching with \texttt{La 2-x Sr x CuO 4}). 
In other cases, the formula was different by formatting, like: \texttt{Nd 2-x Ce x CuO 4} and \texttt{La 2-x Sr x CuO 4}.
However, the more interesting cases were provided by element or amount substitutions such as: \texttt{electron-doped infinite-layer superconductors Sr 0.9 La 0.1 Cu 1-x R x O 2 where R = Zn and Ni} which was matched \texttt{Sr0.9La0.1Cu1-xNixO2}, or \texttt{Eu 1-x K x Fe 2 As 2 samples with x = 0.35, 0.45 and 0.5} and \texttt{Eu 0.5 K 0.5 Fe 2 As 2'}. These two cases were particularly complicated to match because they required a deeper understanding of the formula structure. 

Among the errors of the formula matching, all of them were provided by the formula which was not correctly parsed, for example in one complicated case with the substrate information: \texttt{(1-x/2)La 2 O 3 /xSrCO 3 /CuO in molar ratio with x = 0.063, 0.07, 0.09, 0.10, 0.111 and 0.125} which was incorrectly matched with the general \texttt{La2O3}.

\subsection{NER on properties extraction}
\label{sec:results-ner-properties}

The property extraction assessment was performed using the MeasEval dataset, with the baseline established by Grobid volumes, achieving an approximately 85\% score using a holdout dataset created in conjunction with the application. 
At the time of writing, the evaluation of grobid-quantities~\cite{foppiano2019quantities} (version 0.7.3\footnote{\url{https://github.com/lfoppiano/grobid-quantities/releases/tag/v0.7.3}}) against MeasEval yielded a score of around 59\% F1-score. This disparity was anticipated, given the slightly divergent annotation strategies employed by the MeasEval developers compared to those used in developing grobid-quantities (e.g., considerations such as approximate values and other proximity expressions were not considered). 

\begin{figure}[htbp]
  \centering
  \includegraphics[width=0.78\textwidth]{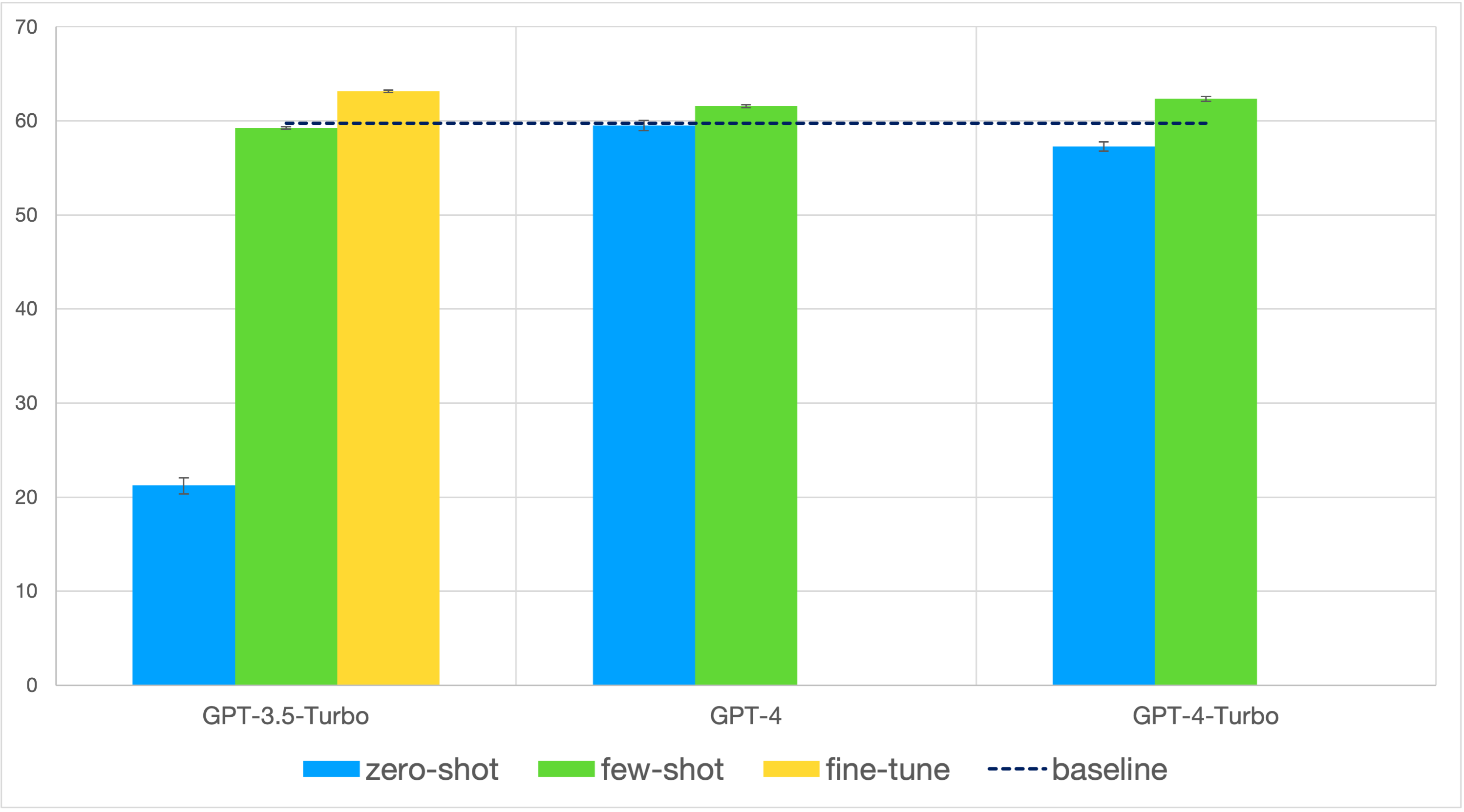} 
  \caption{Comparison scores for properties extraction using NER. The scores are the aggregations of the micro average F1 scores and are calculated using soft matching with a threshold of 0.9 similarity. The error bars are calculated over the standard deviation of three independent runs.}
  \label{fig:ner-measeval-all}
\end{figure}

Unexpectedly, none of the models outperformed grobid-quantities in zero-shot prompting, as depicted in Figure~\ref{fig:ner-measeval-all}. This outcome is surprising considering that a) the expression of properties lacks a specific domain constraint (aside from potential variations in frequency distribution), and b) measurements of physical quantities are likely prevalent in the extensive text corpus used to pre-train the OpenAI models.

In the realm of few-shot prompting (Figure~\ref{fig:ner-measeval-all}), a marginal improvement was observed only for GPT-4 and GPT-4-Turbo, resulting in an F1-score gain ranging around 2\%. 
However, this improvement is not significant. 
We theorise that the hints provided to the LLMs may introduce bias. When these hints are incorrect or incomplete, the LLMs struggle to guide the generation effectively, impacting the quality of the output results.
Significantly, the fine-tuned model (Figure~\ref{fig:ner-measeval-all}) shows a slight enhancement compared to zero-shot, few-shot, and the baseline. Interestingly, in this specific instance where both the baseline and fine-tuned models are trained and evaluated on the same data, the LLM demonstrates an approximate 3\% increase in the F1-score.

\subsection{NER on materials expressions extraction}
\label{sec:results-ner-materials}
The evaluation of material expressions extraction was performed using the partition of the SuperMat~\cite{lfoppiano2021supermat} dataset dedicated to validation, consisting of 32 articles.

\begin{figure}[htbp]
  \centering
  \includegraphics[width=0.8\textwidth]{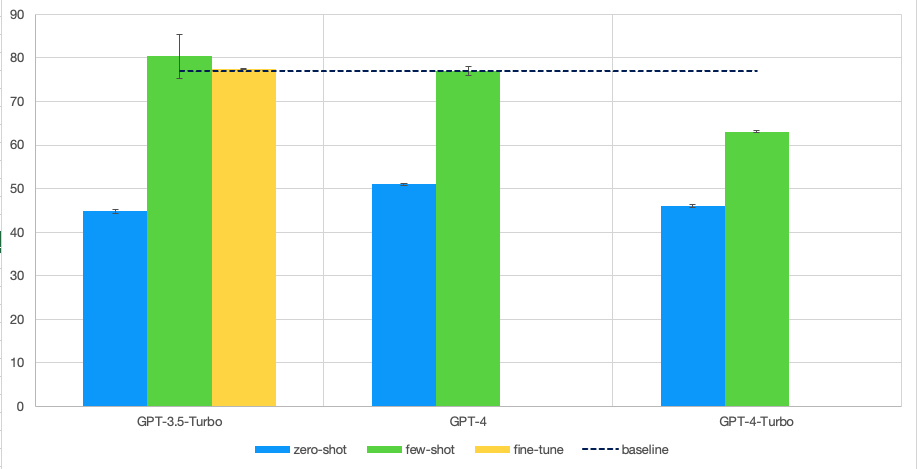} 
  \caption{Comparison scores for material extraction using NER. The metrics are the aggregations of the micro average F1-scores, calculated using formula matching. The error bars are calculated over the standard deviation of three independent runs.}
  \label{fig:ner-materials-all}
\end{figure}

In zero-shot prompting (Figure~\ref{fig:ner-materials-all}), both GPT-4 and GPT-4-Turbo achieved comparable F1-scores, hovering around 50\%. Notably, all LLMs scored at least 10\% lower than the baseline~\cite{lfoppiano2023automatic}. This disparity is expected, given that material expressions may involve extensive sequences and encompass multiple pieces of information not easily conveyed in the prompt.
Few-shot prompting (Figure~\ref{fig:ner-materials-all}) yielded improved results, with GPT-3.5-Turbo and GPT-4 slightly surpassing the baseline. 
The introduction of hints in the prompt indeed enhances performance, but, as previously discussed, it appears to strongly influence the LLMs, not able to mitigate the impact of invalid hints that may be provided.
Equally unexpected, fine-tuning did not outperform few-shot prompting. This outcome suggests that the additional training did not significantly enhance the LLMs' ability to handle material expressions.

\subsection{Relation extraction}
\label{sec:results-re}

\begin{figure}[htbp]
  \centering
  \includegraphics[width=0.8\textwidth]{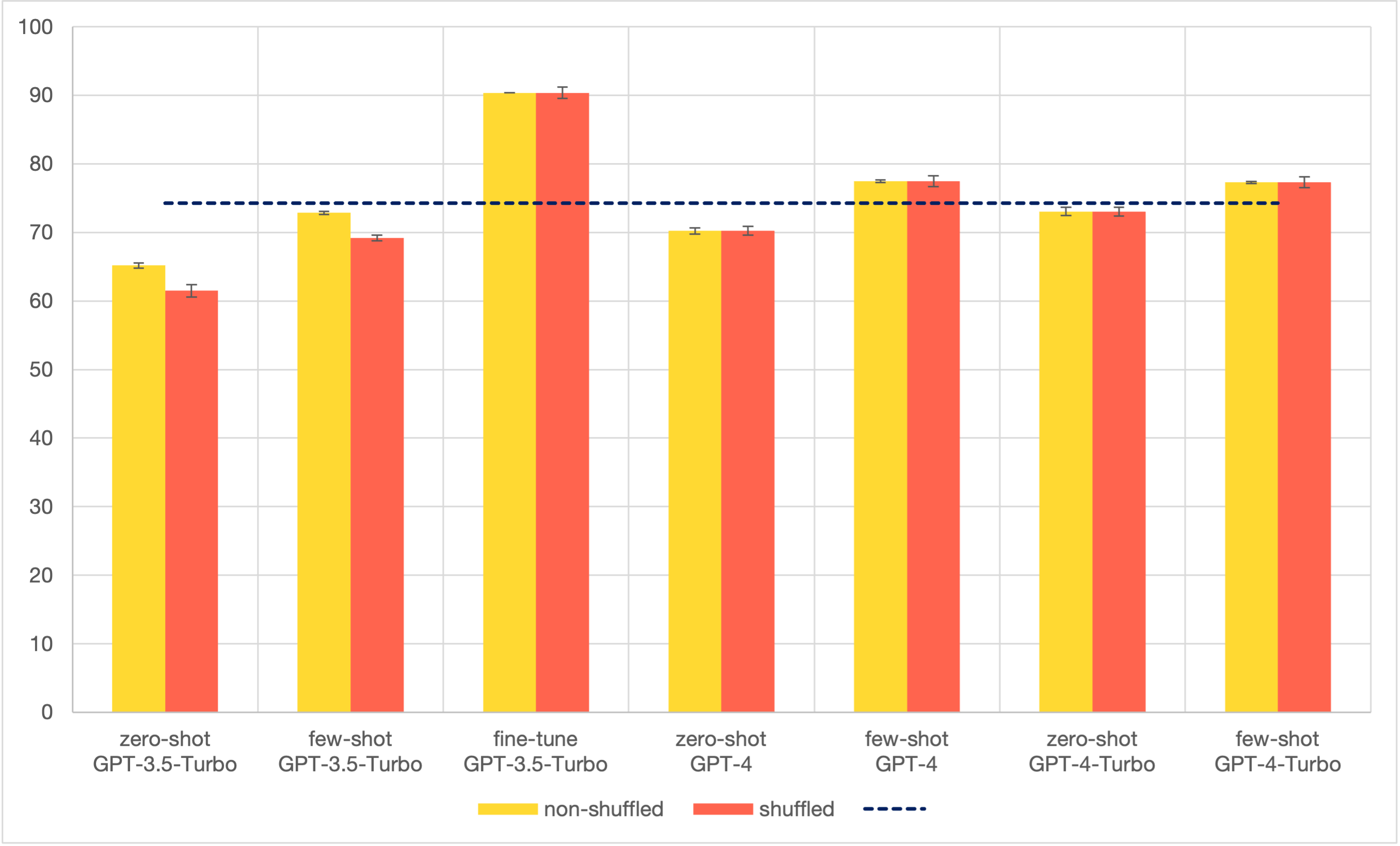} 
  \caption{Comparison of the scores of the shuffled extraction using zero-shot prompting, few-shot prompting and the fine-tuned model for RE on materials and properties. The metrics are the aggregated micro average F1-scores calculated using strict matching. The error bars are calculated over the standard deviation of three independent runs.}
  \label{fig:re-eval-all}
\end{figure}

The evaluation of RE utilised the complete SuperMat dataset, with the results illustrated in Figure~\ref{fig:re-eval-all}, comparing the effects of shuffling across different models. 


GPT-3.5-Turbo zero-shot and few-shot prompting demonstrate a significant difference between shuffled and non-shuffled evaluation (Section~\ref{subsub:shuffled-non_shuffled-eval}), suggesting a sequential connection of entities without specific contextual reasoning.
Notably, the fine-tuned GPT-3.5-Turbo model outperforms the baseline by approximately 15\% F1-score and does not show relevant differences when the evaluation is performed under shuffling conditions.

\begin{figure}[htbp]
  \centering
  \includegraphics[width=0.8\textwidth]{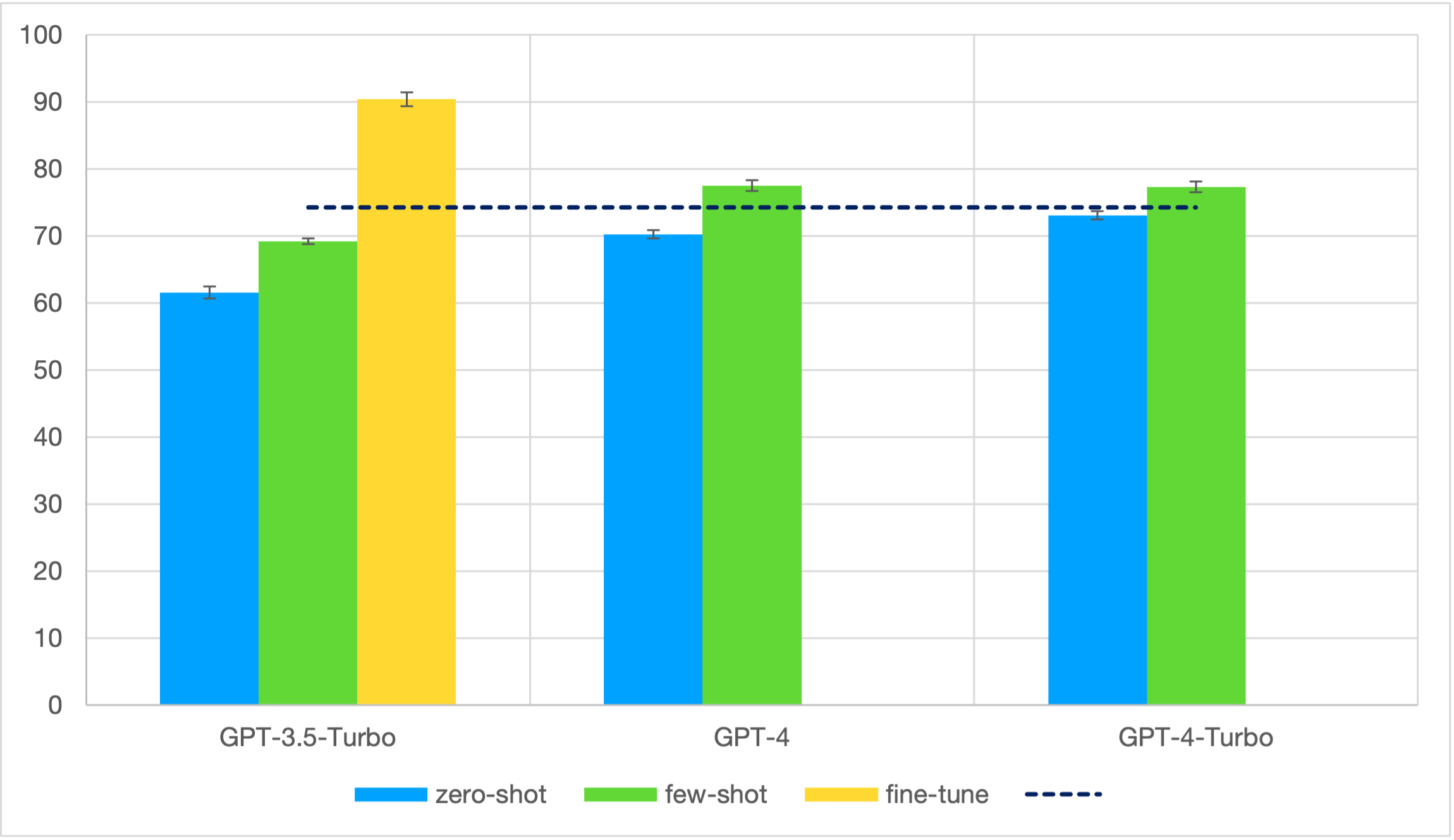} 
  \caption{Overview evaluation on shuffling the provided entities in RE on materials and properties. The metrics are the aggregated micro average F1-scores calculated using strict matching. The error bars are calculated over the standard deviation of three independent runs.}
  \label{fig:re-eval-shuffled-all}
\end{figure}

Figure~\ref{fig:re-eval-shuffled-all} specifically highlights the shuffled version of each model and extraction type. Except for GPT-3.5-Turbo, few-shot prompting shows an improvement compared to zero-shot prompting, achieved by incorporating additional examples in each prompt. 
GPT-4 and GPT-4-Turbo also exhibit stable results under shuffling conditions, achieving an F1-score of around 15-18\% lower than fine-tuned GPT-3.5-Turbo.

\subsubsection{Data variability for fine-tuning}
\label{subsubsec:data-variability}

\begin{figure}[htbp]
  \centering
  \includegraphics[width=0.8\textwidth]{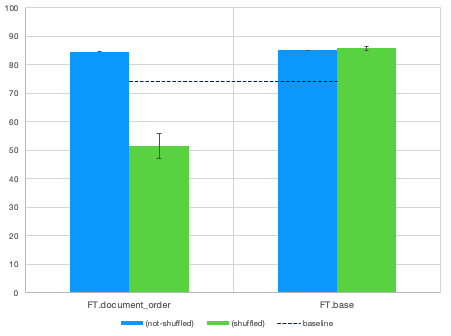} 
  \caption{Evaluation of the impact of data variability in fine-tuning GPT-3.5-Turbo. The metrics are the aggregated micro average F1-scores calculated using strict matching. The model "FT.document\_order" was fine-tuned with the original data, where entities were taken of appearance. In "FT.base", our default strategy, the entities provided to the prompt were scrambled. The error bars are calculated over the standard deviation of three independent runs.}
  \label{fig:re-eval-ft}
\end{figure}

In Section~\ref{subsec:consideration-fine-tuning}, we describe two additional ways to prepare the data for fine-tuning. 
As illustrated in Figure~\ref{fig:re-eval-ft}, the GPT-3.5-Turbo model fine-tuned with the strategy "FT.document\_order" showed an inability to generalise when evaluated under shuffling conditions, where the model loses around 30\% in F1-score.
This suggests that adding entropy (for example, by shuffling the data) should be performed as a best practice, which could result in models with larger reasoning capabilities.

When we increased the size of the dataset used in fine-tuning to almost double (Table~\ref{tab:amount-data-fine-tuned}), the resulting model did not improve compared to the FT.base. These results confirm that in fine-tuning, size does not matter, while data variability and quality do.



\section{Code and data availability}
This work is available at \url{https://github.com/lfoppiano/MatSci-LumEn}. The repository contains the scripts and the data used for extraction and evaluation. 
The code of the material parser used in the formula matching is available at \url{https://github.com/lfoppiano/material-parsers}, and the service API is accessible at \url{https://lfoppiano-material-parsers.hf.space}. 

\section{Conclusion}

In this study, we have proposed an evaluation framework for estimating how well LLMs perform compared with SLMs and rule-based tasks related to materials science by focusing on sub-domains such as superconductor research. The findings obtained from our work provide initial guidance applicable to other materials science sub-domains in future research.

To evaluate material extraction comparison, we proposed a novel method to parse and match formula elements by elements through an aggregated parser for materials. 
This new method provides a more realistic F1 score. Compared with strict matching, we obtained a gain in F1-score from 17\% to 45\% for GPT3.5-Turbo NER at the price of a minimal error rate (2\%). 

We then evaluated LLMs on two tasks: NER for materials and properties and RE for linking them. LLMs underperform significantly on NER tasks than SLMs in material and property extraction (Q1). This finding is particularly surprising considering properties since these expressions are not confined to a specific domain.

In material extraction, GPT-3.5-Turbo with fine-tuning failed to outperform the baseline, and the same holds for any model with few-shot prompting. 
For property extraction, GPT-4 and GPT-4-Turbo with zero-shot prompting perform on par with the baseline. GPT-3.5-Turbo with few-shot and fine-tuning, on the other hand, outperforms the baseline by a marginal increase in points.
Our results suggest that, for material expressions, small specialised models remain the most accurate choice.

The scenario improves for RE (Q2).
With two examples, few-shot prompting demonstrates a significant improvement over the baseline.
GPT-4-Turbo exhibits enhanced reasoning capabilities compared to GPT-4 and GPT-3.5-Turbo.
GPT-3.5-Turbo performs poorly in both zero-shot and few-shot prompting, showing a substantial score decrease when entities are shuffled, which aligns with previous observations.
Nevertheless, fine-tuning yields scores superior to the baseline and other models, showing stability when comparing shuffled and unshuffled evaluations.

In conclusion, to answer Q2, GPT-4 and GPT-4-Turbo showcase effective reasoning capabilities for accurately relating concepts and extracting relations without fine-tuning. However, fine-tuning GPT-3.5-Turbo out yields the best results with a relatively small dataset. 
GPT-4-Turbo, which costs one-third of GPT-4, remains a robust choice given its reasoning capabilities. 
However, for Q1, for extracting complex entities such as materials, we find that training small specialised models remains a more effective approach.

\section*{Acknowledgements}
Our warmest thanks to Patrice Lopez for his continuous support and inspiration with ideas, suggestions, and fruitful discussions.

\section*{Funding}
This work was partially supported by the MEXT Programme: Data Creation and Utilisation-Type Material Research and Development Project (Digital Transformation Initiative Centre for Magnetic Materials) Grant Number JPMXP1122715503.

\section*{Notes on Contributors}
LF developed the scripts for extraction and evaluation and wrote the manuscript.
GL supported the LLM evaluation and implementation and financed access to the OpenAI API. 
GL, TA, and MI reviewed the article.
MI supervised the process and provided the budget.

\bibliography{references}
\bibliographystyle{unsrt}

\section*{Figures \& Tables}

\begin{table}[htbp]
    \centering
    \caption{Datasets and support information for fine-tuning GPT-3.5-Turbo. For each task, the data was divided into 70/30 partitions for training and testing, respectively. The testing dataset is different from the evaluation dataset.  }
    \label{tab:amount-data-fine-tuned}
    \begin{tabular}{lcccc}
        \textbf{Task} & \textbf{Preparation strategy} & \textbf{Dataset} & \textbf{\# Training} & \textbf{\# Test} \\
        \toprule
        NER & N/A & SuperMat   & 1639 & 703 \\
        NER & N/A & grobid-quantities dataset & 485 & 208 \\
        RE  & FT.base/FT.document & SuperMat   & 344 & 148 \\
        RE  & FT.augmented & SuperMat & 695 & 299 \\
        \bottomrule
    \end{tabular}
\end{table}

\clearpage

\appendix
\label{appendix}

\section{Full evaluation results}
\label{appendix:full-evaluation-results}

\subsection{NER for properties extraction}

\subsubsection{Zero-shot prompting}
\begin{table}[htbp]
\small
  \centering
  \caption{Performance Metrics for GPT-3.5-turbo NER in properties extraction, zero-shot prompting. P: Precision, R: Recall, F1: harmonic average of P and R, Supp: Support, number of extracted entities.}
  \begin{tabular}{lccccc}
    \toprule
    \textbf{Run} & \textbf{Matching} & \textbf{P} & \textbf{R} & \textbf{F1} & \textbf{Supp} \\
    \midrule
    \multirow{2}{*}{Run1} & Soft matching & 42.73 & 14.49 & 21.64 & 564 \\
    & Sentence BERT & 45.39 & 15.39 & 22.99 & 564 \\
    \midrule
    \multirow{2}{*}{Run2} & Soft matching & 41.81 & 13.35 & 20.24 & 531 \\
    & Sentence BERT & 45.01 & 14.37 & 21.79 & 531 \\
    \midrule
    \multirow{2}{*}{Run3} & Soft matching & 42.86 & 14.61 & 21.79 & 567 \\
    & Sentence BERT & 45.86 & 15.63 & 23.32 & 567 \\
    \midrule
    \multicolumn{2}{l}{\textbf{Mean and Standard deviation of F1 score}} & & & & \\
    \midrule
    \textbf{Matching} & \textbf{Avg.} & $\sigma\textsuperscript{2}$ & & & \textbf{Avg. Supp}\\
    Soft matching & 21.22 & 0.85 & & & 554 \\
    Sentence BERT & 22.7 & 0.80 & & & \\
    \bottomrule
  \end{tabular}
\end{table}

\begin{table}[htbp]
\small
  \centering
  \caption{Performance Metrics for GPT-4 NER in properties extraction, zero-shot prompting. P: Precision, R: Recall, F1: harmonic average of P and R, Supp: Support, number of extracted entities.}
  \begin{tabular}{lccccc}
    \toprule
    \textbf{Run} & \textbf{Matching} & \textbf{P} & \textbf{R} & \textbf{F1} & \textbf{Supp} \\
    \midrule
    \multirow{2}{*}{Run1} & Soft matching & 61.43 & 56.70 & 58.97 & 1535 \\
    & Sentence BERT & 65.08 & 60.07 & 62.48 & 1535 \\
    \midrule
    \multirow{2}{*}{Run2} & Soft matching & 63.23 & 57.19 & 60.06 & 1504 \\
    & Sentence BERT & 66.42 & 60.07 & 63.09 & 1504 \\
    \midrule
    \multirow{2}{*}{Run3} & Soft matching & 62.83 & 56.52 & 59.51 & 1496 \\
    & Sentence BERT & 66.11 & 59.47 & 62.61 & 1496 \\
    \midrule
    \multicolumn{2}{l}{\textbf{Mean and Standard deviation of F1 score}} & & & & \\
    \midrule
    \textbf{Matching} & \textbf{Avg.} & $\sigma\textsuperscript{2}$ & & & \textbf{Avg. Supp}\\
    Soft matching & 59.51 & 0.54 & & & 1511 \\
    Sentence BERT & 62.72 & 0.32 & & & \\
    \bottomrule
  \end{tabular}
\end{table}

\begin{table}[htbp]
\small
  \centering
  \caption{Performance Metrics for GPT4-turbo NER in properties extraction, zero-shot prompting. P: Precision, R: Recall, F1: harmonic average of P and R, Supp: Support, number of extracted entities.}
  \begin{tabular}{lccccc}
    \toprule
    \textbf{Run} & \textbf{Matching} & \textbf{P} & \textbf{R} & \textbf{F1} & \textbf{Supp} \\
    \midrule
    \multirow{2}{*}{Run1} & Soft matching & 59.60 & 54.30 & 56.83 & 1515 \\
    & Sentence BERT & 62.90 & 57.31 & 59.97 & 1515 \\
    \midrule        
    \multirow{2}{*}{Run2} & Soft matching & 60.34 & 55.44 & 57.79 & 1528 \\
    & Sentence BERT & 63.87 & 58.69 & 61.17 & 1528 \\
    \midrule        
    \multirow{2}{*}{Run3} & Soft matching & 60.05 & 54.60 & 57.20 & 1512 \\    
    & Sentence BERT & 63.96 & 58.15 & 60.91 & 1512 \\
    \midrule
    \multicolumn{2}{l}{\textbf{Mean and Standard deviation of F1 score}} & & & & \\
    \midrule
    \textbf{Matching} & \textbf{Avg.} & $\sigma\textsuperscript{2}$ & & & \textbf{Avg. Supp}\\
    Soft matching & 57.27 & 0.48 & & & 1518 \\
    Sentence BERT & 60.68 & 0.63 & & & \\
    \bottomrule
  \end{tabular}
\end{table}

\clearpage
\subsubsection{Few-shot prompting }

\begin{table}[htbp]
\small
  \centering
  \caption{Performance Metrics for GPT-3.5-turbo NER in properties extraction, few-shot prompting. P: Precision, R: Recall, F1: harmonic average of P and R, Supp: Support, number of extracted entities.}
  \begin{tabular}{lccccc}
    \toprule
    \textbf{Run} & \textbf{Matching} & \textbf{P} & \textbf{R} & \textbf{F1} & \textbf{Supp} \\
    \midrule
    \multirow{2}{*}{Run1} & Soft matching & 60.98 & 57.79 & 59.34 & 1576 \\
    & Sentence BERT & 64.85 & 61.46 & 63.11 & 1576 \\    
    \midrule
    \multirow{2}{*}{Run2} & Soft matching & 60.99 & 57.73 & 59.31 & 1574 \\
    & Sentence BERT & 64.55 & 61.09 & 62.77 & 1574 \\
    \midrule
    \multirow{2}{*}{Run3} & Soft matching & 60.72 & 57.55 & 59.09 & 1576 \\
    & Sentence BERT & 64.28 & 60.91 & 62.55 & 1576 \\
    \midrule
    \multicolumn{2}{l}{\textbf{Mean and Standard deviation of F1 score}} & & & & \\
    \midrule
    \textbf{Matching} & \textbf{Avg.} & $\sigma\textsuperscript{2}$ & & & \textbf{Avg. Supp}\\
    Soft matching & 59.24 & 0.13 & & & 1575 \\
    Sentence BERT & 62.81 & 0.28 & & & \\
    \bottomrule
  \end{tabular}
\end{table}

\begin{table}[htbp]
\small
  \centering
  \caption{Performance Metrics for GPT-4 NER in properties extraction, few-shot prompting. P: Precision, R: Recall, F1: harmonic average of P and R, Supp: Support, number of extracted entities.}
  \begin{tabular}{lccccc}
    \toprule
    \textbf{Run} & \textbf{Matching} & \textbf{P} & \textbf{R} & \textbf{F1} & \textbf{Supp} \\
    \midrule
    \multirow{2}{*}{Run1} & Soft matching & 62.01 & 60.85 & 61.43 & 1632 \\
    & Sentence BERT & 65.56 & 64.34 & 64.95 & 1632 \\
    \midrule    
    \multirow{2}{*}{Run2} & Soft matching & 62.15 & 60.91 & 61.52 & 1630 \\
    & Sentence BERT & 65.52 & 64.22 & 64.86 & 1630 \\
    \midrule    
    \multirow{2}{*}{Run3} & Soft matching & 62.41 & 61.09 & 61.74 & 1628 \\
    & Sentence BERT & 65.6 & 64.22 & 64.9 & 1628 \\
    \midrule    
    \multicolumn{2}{l}{\textbf{Mean and Standard deviation of F1 score}} & & & & \\
    \midrule
    \textbf{Matching} & \textbf{Avg.} & $\sigma\textsuperscript{2}$ & & & \textbf{Avg. Supp}\\
    Soft matching & 61.56 & 0.16 & & & 1630 \\
    Sentence BERT & 64.9 & 0.04 & & & \\
    \bottomrule
  \end{tabular}
\end{table}

\begin{table}[htbp]
\small
  \centering
  \caption{Performance Metrics for GPT-4-turbo NER in properties extraction, few-shot prompting. P: Precision, R: Recall, F1: harmonic average of P and R, Supp: Support, number of extracted entities.}
  \begin{tabular}{lccccc}
    \toprule
    \textbf{Run} & \textbf{Matching} & \textbf{P} & \textbf{R} & \textbf{F1} & \textbf{Supp} \\
    \midrule
    \multirow{2}{*}{Run1} & Soft matching & 62.18 & 62 & 62.09 & 1652 \\
    & Sentence BERT & 65.8 & 65.6 & 65.7 & 1652 \\
    \midrule    
    \multirow{2}{*}{Run2} & Soft matching & 62.42 & 62.24 & 62.33 & 1658 \\
    & Sentence BERT & 66.1 & 65.9 & 66 & 1658 \\
    \midrule    
    \multirow{2}{*}{Run3} & Soft matching & 62.48 & 62.78 & 62.63 & 1671 \\
    & Sentence BERT & 65.95 & 66.27 & 66.11 & 1671 \\
    \midrule
    \multicolumn{2}{l}{\textbf{Mean and Standard deviation of F1 score}} & & & & \\
    \midrule
    \textbf{Matching} & \textbf{Avg.} & $\sigma\textsuperscript{2}$ & & & \textbf{Avg. Supp}\\
    Soft matching & 62.35 & 0.27 & & & 1660 \\
    Sentence BERT & 65.93 & 0.21 & & & \\
    \bottomrule
  \end{tabular}
\end{table}

\clearpage
\subsubsection{Fine-tuning}

\begin{table}[htbp]
\small
  \centering
  \caption{Performance Metrics for fine-tuned GPT-3.5-turbo NER in properties extraction. P: Precision, R: Recall, F1: harmonic average of P and R, Supp: Support, number of extracted entities.}
  \begin{tabular}{lccccc}
    \toprule
    \textbf{Run} & \textbf{Matching} & \textbf{P} & \textbf{R} & \textbf{F1} & \textbf{Supp} \\
    \midrule
    \multirow{2}{*}{Run1} & Soft matching & 60.76 & 58.57 & 59.64 & 1603 \\
    & Sentence BERT & 64.38 & 62.06 & 63.2 & 1603 \\
    \midrule
    \multirow{2}{*}{Run2} & Soft matching & 60.85 & 58.69 & 59.75 & 1604 \\
    & Sentence BERT & 64.4 & 62.12 & 63.24 & 1604 \\
    \midrule
    \multirow{2}{*}{Run3} & Soft matching & 60.49 & 58.45 & 59.45 & 1607 \\
    & Sentence BERT & 64.09 & 61.94 & 63 & 1607 \\
    \midrule
    \multicolumn{2}{l}{\textbf{Mean and Standard deviation of F1 score}} & & & & \\
    \midrule
    \textbf{Matching} & \textbf{Avg.} & $\sigma\textsuperscript{2}$ & & & \textbf{Avg. Supp}\\
    Soft matching & 59.61 & 0.15 & & & 1604 \\
    Sentence BERT & 63.14 & 0.12 & & & \\
    \bottomrule
  \end{tabular}
\end{table}

\clearpage
\subsection{NER for Materials extraction}

\subsubsection{Zero-shot}

\begin{table}[htbp]
\small
  \centering
  \caption{Performance Metrics for GPT3.5-turbo NER in materials extraction, zero-shot prompting. P: Precision, R: Recall, F1: harmonic average of P and R, Supp: Support, number of extracted entities.}
  \begin{tabular}{lccccc}
    \toprule
    \textbf{Run} & \textbf{Matching} & \textbf{P} & \textbf{R} & \textbf{F1} & \textbf{Supp} \\
    \midrule
    \multirow{2}{*}{Run1} & Sentence BERT & 37.34 & 22.58 & 28.14 & 1617 \\
    & Formula & 59.49 & 35.97 & 44.83 & 1617 \\
    \midrule
    \multirow{2}{*}{Run2} & Sentence BERT & 37.06 & 22.83 & 28.26 & 1641 \\
    & Formula & 59.21 & 36.48 & 45.15 & 1641 \\
    \midrule
    \multirow{2}{*}{Run3} & Sentence BERT & 37.45 & 22.07 & 27.77 & 1587 \\
    & Formula & 59.74 & 35.2 & 44.3 & 1587 \\
    \midrule
    \multicolumn{2}{l}{\textbf{Mean and Standard deviation of F1 score}} & & & & \\
    \midrule
    \textbf{Matching} & \textbf{Avg.} & $\sigma\textsuperscript{2}$ & & & \textbf{Avg. Supp}\\
    Sentence BERT & 28.05 & 0.25 & & & 1615 \\
    Formula & 44.76 & 0.42 & & & \\
    \bottomrule
  \end{tabular}
\end{table}

\begin{table}[htbp]
\small
  \centering
  \caption{Performance Metrics for GPT-4 NER in materials extraction, zero-shot prompting. P: Precision, R: Recall, F1: harmonic average of P and R, Supp: Support, number of extracted entities.}
  \begin{tabular}{lccccc}
    \toprule
    \textbf{Run} & \textbf{Matching} & \textbf{P} & \textbf{R} & \textbf{F1} & \textbf{Supp} \\
    \midrule
    \multirow{2}{*}{Run1} & Sentence BERT & 49.9 & 31.25 & 38.43 & 1103 \\
    & Formula & 66.4 & 41.58 & 51.14& 1103 \\
    \midrule
    \multirow{2}{*}{Run2} & Sentence BERT & 49.9 & 30.61 & 37.94 & 1097 \\
    & Formula & 66.94 & 41.07 & 50.91 & 1097 \\
    \midrule
    \multirow{2}{*}{Run3} & Sentence BERT & 49.59 & 30.74 & 37.95 & 1108 \\
    & Formula & 66.46 & 41.2 & 50.87 & 1108 \\
    \midrule
    \multicolumn{2}{l}{\textbf{Mean and Standard deviation of F1 score}} & & & & \\
    \midrule
    \textbf{Matching} & \textbf{Avg.} & $\sigma\textsuperscript{2}$ & & & \textbf{Avg. Supp}\\
    Sentence BERT & 38.10 & 0.28 & & & 1102 \\
    Formula & 50.97 & 0.14 & & & \\
    \bottomrule
  \end{tabular}
\end{table}

\begin{table}[htbp]
\small
  \centering
  \caption{Performance Metrics for GPT4-turbo NER in properties extraction, zero-shot prompting. P: Precision, R: Recall, F1: harmonic average of P and R, Supp: Support, number of extracted entities.}
  \begin{tabular}{lccccc}
    \toprule
    \textbf{Run} & \textbf{Matching} & \textbf{P} & \textbf{R} & \textbf{F1} & \textbf{Supp} \\
    \midrule
    \multirow{2}{*}{Run1} & Sentence BERT & 45.53 & 22.07 & 29.73 & 883 \\
    & Formula & 70.53 & 34.18 & 46.05 & 883 \\
    \midrule
    \multirow{2}{*}{Run2} & Sentence BERT & 46.67 & 22.32 & 30.2 & 873 \\
    & Formula & 70.67 & 33.8 & 45.73 & 873 \\
    \midrule
    \multirow{2}{*}{Run3} & Sentence BERT & 46.74 & 22.83 & 30.68 & 878 \\
    & Formula & 70.5 & 34.44 & 46.27 & 878 \\
    \midrule
    \multicolumn{2}{l}{\textbf{Mean and Standard deviation of F1 score}} & & & & \\
    \midrule
    \textbf{Matching} & \textbf{Avg.} & $\sigma\textsuperscript{2}$ & & & \textbf{Avg. Supp}\\
    Sentence BERT & 45.98 & 0.56 & & & 878 \\
    Formula & 53.60 & 0.30 & & & \\
    \bottomrule
  \end{tabular}
\end{table}

\clearpage
\subsubsection{Few-shot}

\begin{table}[htbp]
\small
  \centering
  \caption{Performance Metrics for GPT3.5-turbo NER in materials extraction, few-shot prompting. P: Precision, R: Recall, F1: harmonic average of P and R, Supp: Support, number of extracted entities.}
  \begin{tabular}{lccccc}
    \toprule
    \textbf{Run} & \textbf{Matching} & \textbf{P} & \textbf{R} & \textbf{F1} & \textbf{Supp} \\
    \midrule
    \multirow{2}{*}{Run1} & Sentence BERT & 71.58 & 69.39 & 70.47 & 1887 \\
    & Formula & 75.79 & 73.47 & 74.61 & 1887 \\
    \midrule
    \multirow{2}{*}{Run2} & Sentence BERT & 78.73 & 88.78 & 83.45 & 2495 \\
    & Formula & 78.39 & 88.39 & 83.09 & 2495 \\
    \midrule
    \multirow{2}{*}{Run3} & Sentence BERT & 78.85 & 88.9 & 83.57 & 2448 \\
    & Formula & 78.73 & 88.78 & 83.45 & 2448 \\
    \midrule
    \multicolumn{2}{l}{\textbf{Mean and Standard deviation of F1 score}} & & & & \\
    \midrule
    \textbf{Matching} & \textbf{Avg.} & $\sigma\textsuperscript{2}$ & & & \textbf{Avg. Supp}\\
    Sentence BERT & 79.16 & 7.5 & & & 2276 \\
    Formula & 80.38 & 4.82 & & & \\
    \bottomrule
  \end{tabular}
\end{table}

\begin{table}[htbp]
\small
  \centering
  \caption{Performance Metrics for GPT-4 NER in materials extraction, few-shot prompting. P: Precision, R: Recall, F1: harmonic average of P and R, Supp: Support, number of extracted entities.}
  \begin{tabular}{lccccc}
    \toprule
    \textbf{Run} & \textbf{Matching} & \textbf{P} & \textbf{R} & \textbf{F1} & \textbf{Supp} \\
    \midrule
    \multirow{2}{*}{Run1} & Sentence BERT & 75.87 & 75 & 75.43 & 1402 \\
    & Formula & 77.16 & 76.28 & 76.72 & 1412 \\
    \midrule
    \multirow{2}{*}{Run2} & Sentence BERT & 70.67 & 82.65 & 76.19 & 1402 \\
    & Formula & 70.99 & 83.04 & 76.54 & 1854 \\
    \midrule
    \multirow{2}{*}{Run3} & Sentence BERT & 70.99 & 83.04 & 76.54 & 1402 \\
    & Formula & 72.09 & 83.67 & 77.45 & 1826 \\
    \midrule
    \multicolumn{2}{l}{\textbf{Mean and Standard deviation of F1 score}} & & & & \\
    \midrule
    \textbf{Matching} & \textbf{Avg.} & $\sigma\textsuperscript{2}$ & & & \textbf{Avg. Supp}\\
    Sentence BERT & 76.35 & 1.02 & & & 1402 \\
    Formula & 76.90 & 0.48 & & & \\
    \bottomrule
  \end{tabular}
\end{table}

\begin{table}[htbp]
\small
  \centering
  \caption{Performance Metrics for GPT4-turbo NER in materials extraction, few-shot prompting. P: Precision, R: Recall, F1: harmonic average of P and R, Supp: Support, number of extracted entities.}
  \begin{tabular}{lccccc}
    \toprule
    \textbf{Run} & \textbf{Matching} & \textbf{P} & \textbf{R} & \textbf{F1} & \textbf{Supp} \\
    \midrule
    \multirow{2}{*}{Run1} & Sentence BERT & 54.78 & 59.95 & 57.25 & 1735 \\
    & Formula & 60.26 & 65.94 & 62.97 & 1735 \\
    \midrule
    \multirow{2}{*}{Run2} & Sentence BERT & 55.54 & 59.44 & 57.42 & 1707 \\
    & Formula & 60.91 & 65.18 & 62.97 & 1707 \\
    \midrule
    \multirow{2}{*}{Run3} & Sentence BERT & 55.81 & 60.08 & 57.86 & 1707 \\
    & Formula & 61.02 & 65.69 & 63.27 & 1707 \\
    \midrule
    \multicolumn{2}{l}{\textbf{Mean and Standard deviation of F1 score}} & & & & \\
    \midrule
    \textbf{Matching} & \textbf{Avg.} & $\sigma\textsuperscript{2}$ & & & \textbf{Avg. Supp}\\
    Sentence BERT & 57.51 & 0.31 & & & 1716 \\
    Formula & 63.07 & 0.17 & & & \\
    \bottomrule
  \end{tabular}
\end{table}

\clearpage
\subsubsection{Fine-tuning}

\begin{table}[htbp]
\small
  \centering
  \caption{Performance Metrics for the fine-tuned GPT3.5-turbo NER in materials extraction. P: Precision, R: Recall, F1: harmonic average of P and R, Supp: Support, number of extracted entities.}
  \begin{tabular}{lccccc}
    \toprule
    \textbf{Run} & \textbf{Matching} & \textbf{P} & \textbf{R} & \textbf{F1} & \textbf{Supp} \\
    \midrule
    \multirow{2}{*}{Run1} & Sentence BERT & 61.02 & 65.69 & 63.27 & 1429 \\
    & Formula & 61.02 & 65.69 & 63.27 & 1429 \\
    \midrule
    \multirow{2}{*}{Run2} & Sentence BERT & 72.24 & 67.73 & 69.91 & 1429 \\
    & Formula & 80.14 & 75.13 & 77.55 & 1429 \\
    \midrule
    \multirow{2}{*}{Run3} & Sentence BERT & 72.17 & 67.47 & 69.74 & 1432 \\
    & Formula & 80.08 & 74.87 & 77.39 & 1432 \\
    \midrule
    \multicolumn{2}{l}{\textbf{Mean and Standard deviation of F1 score}} & & & & \\
    \midrule
    \textbf{Matching} & \textbf{Avg.} & $\sigma\textsuperscript{2}$ & & & \textbf{Avg. Supp}\\
    Sentence BERT & 69.75 & 0.15 & & & 1430 \\
    Formula & 77.43 & 0.09 & & & \\
    \bottomrule
  \end{tabular}
\end{table}

\clearpage
\subsection{RE for Materials-Properties extraction}

\subsubsection{Zero-shot}

\begin{table}[htbp]
  \small
  \centering
  \caption{Performance Metrics for the GPT3.5-turbo model on RE in materials-properties extraction. In this run, the entities are used as they appear in the input data when included in the prompt. P: Precision, R: Recall, F1: harmonic average of P and R, Supp: Support, number of extracted entities.}
  \begin{tabular}{lccccc}
    \toprule
    \textbf{Run} & \textbf{Matching} & \textbf{P} & \textbf{R} & \textbf{F1} & \textbf{Supp} \\
    \midrule
    \multirow{2}{*}{Run1} & Strict matching & 79.90 & 55.29 & 65.36 & 791 \\
    & Soft matching & 80.15 & 55.47 & 65.56 & 791 \\
    \midrule
    \multirow{2}{*}{Run2} & Strict matching & 80.13 & 54.33 & 64.75 & 775 \\
    & Soft matching & 80.52 & 54.59 & 65.07 & 775 \\
    \midrule
    \multirow{2}{*}{Run3} & Strict matching & 80.74 & 55.03 & 65.45 & 779 \\
    & Soft matching & 81.00 & 55.21 & 65.66 & 779 \\
    \midrule
    \multicolumn{2}{l}{\textbf{Mean and Standard deviation of F1 score}} & & & & \\
    \midrule
    \textbf{Matching} & \textbf{Avg.} & $\sigma\textsuperscript{2}$ & & & \textbf{Avg. Supp}\\
    Strict matching & 65.18 & 0.38 & & & 781 \\
    Soft matching & 65.43 & 0.31 & & & \\
    \bottomrule
  \end{tabular}
\end{table}

\begin{table}[htbp]
    \small
    \centering
    \caption{Performance Metrics for the GPT3.5-turbo model on RE in materials-properties extraction. In this run, we shuffle entities to make it more challenging for the LLM. P: Precision, R: Recall, F1: harmonic average of P and R, Supp: Support, number of extracted entities.}
    \begin{tabular}{lccccc}
        \toprule
        \textbf{Run} & \textbf{Matching} & \textbf{P} & \textbf{R} & \textbf{F1} & \textbf{Supp} \\
        \midrule
        \multirow{2}{*}{Run1} & Strict matching & 76.01 & 51.27 & 61.23 & 771 \\
        & Soft matching & 76.91 & 51.88 & 61.96 & 771  \\
        \midrule
        \multirow{2}{*}{Run2} & Strict matching & 75.82 & 50.74 & 60.80 & 765 \\
        & Soft matching & 76.34 & 51.09 & 61.22 & 765 \\
        \midrule
        \multirow{2}{*}{Run3} & Strict matching & 76.55 & 52.84 & 62.53 & 789 \\
        & Soft matching & 77.19 & 53.28 & 63.04 & 789  \\
        \midrule
        \multicolumn{2}{l}{\textbf{Mean and Standard deviation of F1 score}} & & & & \\
        \midrule
        \textbf{Matching} & \textbf{Avg.} & $\sigma\textsuperscript{2}$ & & & \textbf{Avg. Supp}\\
        Strict matching & 61.52 & 0.90 & & & 775 \\
        Soft matching & 62.07 & 0.91 & & \\
        \bottomrule
    \end{tabular}
\end{table}


\begin{table}[htbp]
    \small
    \centering
    \caption{Performance Metrics for the GPT-4 model on RE in materials-properties extraction. In this run, entities are not shuffled when added to the prompt. P: Precision, R: Recall, F1: harmonic average of P and R, Supp: Support, number of extracted entities.}
    \begin{tabular}{lccccc}
        \toprule
        \textbf{Run} & \textbf{Matching} & \textbf{P} & \textbf{R} & \textbf{F1} & \textbf{Supp} \\
        \midrule
        \multirow{2}{*}{Run1} & Strict matching & 76.87 & 66.58 & 71.35 & 990 \\
        & Soft matching & 77.47 & 67.1 & 71.92 & 990 \\
        \midrule
        \multirow{2}{*}{Run2} & Strict matching & 76.4 & 67.98 & 71.94 & 1017 \\
        & Soft matching & 76.5 & 68.07 & 72.04 & 1017 \\
        \midrule
        \multirow{2}{*}{Run3} & Strict matching & 76.8 & 68.07 & 72.17 & 1013 \\
        & Soft matching & 76.9 & 68.15 & 72.26 & 1013 \\
        \midrule
        \multicolumn{2}{l}{\textbf{Mean and Standard deviation of F1 score}} & & & & \\
        \midrule
        \textbf{Matching} & \textbf{Avg.} & $\sigma\textsuperscript{2}$ & & & \textbf{Avg. Supp}\\
        Strict matching & 71.82 & 0.42 & & & 1006 \\
        Soft matching   & 72.07 & 0.17 & & \\
        \bottomrule
    \end{tabular}
\end{table}

\begin{table}[htbp]
    \small
    \centering
    \caption{Performance Metrics for the GPT4 model on RE in materials-properties extraction. In this run, entities are shuffled to make it more challenging for the LLM. P: Precision, R: Recall, F1: harmonic average of P and R, Supp: Support, number of extracted entities.}
    \begin{tabular}{lccccc}
        \toprule
        \textbf{Run} & \textbf{Matching} & \textbf{P} & \textbf{R} & \textbf{F1} & \textbf{Supp} \\
        \midrule
        \multirow{2}{*}{Run1} & Strict matching & 75.86 & 65.97 & 70.57 & 994 \\
        & Soft matching & 76.16 & 66.23 & 70.85 & 994 \\
        \midrule
        \multirow{2}{*}{Run2} & Strict matching & 74.06 & 65.44 & 69.48 & 1010 \\
        & Soft matching & 74.06 & 65.44 & 69.48 & 1010 \\
        \midrule
        \multirow{2}{*}{Run3} & Strict matching & 75.37 & 66.4 & 70.6 & 1007 \\
        & Soft matching & 75.57 & 66.58 & 70.79 & 1007 \\
        \midrule
        \multicolumn{2}{l}{\textbf{Mean and Standard deviation of F1 score}} & & & & \\
        \midrule
        \textbf{Matching} & \textbf{Avg.} & $\sigma\textsuperscript{2}$ & & & \textbf{Avg. Supp}\\
        Strict matching & 70.21 & 0.63 & & & 1003 \\
        Soft matching   & 70.37 & 0.77 & & \\
        \bottomrule
    \end{tabular}
\end{table}

\begin{table}[htbp]
    \small
    \centering
    \caption{Performance Metrics for the GPT4-turbo model on RE in materials-properties extraction. In this run, entities are not shuffled when added to the prompt. P: Precision, R: Recall, F1: harmonic average of P and R, Supp: Support, number of extracted entities.}
    \begin{tabular}{lccccc}
        \toprule
        \textbf{Run} & \textbf{Matching} & \textbf{P} & \textbf{R} & \textbf{F1} & \textbf{Supp} \\
        \midrule
        \multirow{2}{*}{Run1} & Strict matching & 79.8 & 69.12 & 74.07 & 990 \\
        & Soft matching & 80.61 & 69.82 & 74.82 & 990 \\
        \midrule
        \multirow{2}{*}{Run2} & Strict matching & 80.22 & 69.2 & 74.31 & 986 \\
        & Soft matching & 81.03 & 69.9 & 75.06 & 986 \\
        \midrule
        \multirow{2}{*}{Run3} & Strict matching & 78.87 & 73.65 & 76.17 & 989 \\
        & Soft matching & 79.88 & 69.12 & 74.11 & 989 \\
        \midrule
        \multicolumn{2}{l}{\textbf{Mean and Standard deviation of F1 score}} & & & & \\
        \midrule
        \textbf{Matching} & \textbf{Avg.} & $\sigma\textsuperscript{2}$ & & & \textbf{Avg. Supp}\\
        Strict matching & 73.85	& 0.60 & & & 988 \\
        Soft matching & 74.66 & 0.49 & & \\
        \bottomrule
    \end{tabular}
\end{table}

\begin{table}[htbp]
    \small
    \centering
    \caption{Performance Metrics for the GPT4-turbo model on RE in materials-properties extraction. In this run, entities are shuffled to make it more challenging for the LLM. P: Precision, R: Recall, F1: harmonic average of P and R, Supp: Support, number of extracted entities.}
    \begin{tabular}{lccccc}
        \toprule
        \textbf{Run} & \textbf{Matching} & \textbf{P} & \textbf{R} & \textbf{F1} & \textbf{Supp} \\
        \midrule
        \multirow{2}{*}{Run1} & Strict matching & 80.04 & 67.37 & 73.16 & 962 \\
        & Soft matching & 80.98 & 68.15 & 74.01 & 962 \\
        \midrule
        \multirow{2}{*}{Run2} & Strict matching & 80.71 & 67.72 & 73.64 & 959 \\
        & Soft matching & 81.44 & 68.33 & 74.31 & 959 \\
        \midrule
        \multirow{2}{*}{Run3} & Strict matching & 79.27 & 66.58 & 72.37 & 960 \\
        & Soft matching & 80.21 & 66.58 & 72.37 & 960 \\
        \midrule
        \multicolumn{2}{l}{\textbf{Mean and Standard deviation of F1 score}} & & & & \\
        \midrule
        \textbf{Matching} & \textbf{Avg.} & $\sigma\textsuperscript{2}$ & & & \\
        Strict matching & 76.05 & 0.74 & & & 960 \\
        Soft matching & 76.87 & 0.64 & & & \\
        \bottomrule
    \end{tabular}
\end{table}

\clearpage
\subsubsection{Few-shot}


\begin{table}[htbp]
    \small
    \centering
    \caption{Performance Metrics for the GPT3.5-turbo model on RE in materials-properties extraction. In this run, entities are shuffled to make it more challenging for the LLM. P: Precision, R: Recall, F1: harmonic average of P and R, Supp: Support, number of extracted entities.}
    \begin{tabular}{lccccc}
        \toprule
        \textbf{Run} & \textbf{Matching} & \textbf{P} & \textbf{R} & \textbf{F1} & \textbf{Supp} \\
        \midrule
        \multirow{2}{*}{Run1} & Strict matching & 76.38 & 70.17 & 73.14 & 1050 \\
        & Soft matching & 76.95 & 70.69 & 73.69 & 1050 \\
        \midrule
        \multirow{2}{*}{Run2} & Strict matching & 75.78 & 69.82 & 72.68 & 1053 \\
        & Soft matching & 76.35 & 70.34 & 73.22 & 1053 \\
        \midrule
        \multirow{2}{*}{Run3} & Strict matching & 75.93 & 69.82 & 72.74 & 1051 \\
        & Soft matching & 76.78 & 70.6 & 73.56 & 1051 \\
        \midrule
        \multicolumn{2}{l}{\textbf{Mean and Standard deviation of F1 score}} & & & & \\
        \midrule
        \textbf{Matching} & \textbf{Avg.} & $\sigma\textsuperscript{2}$ & & & \textbf{Avg. Supp}\\
        Strict matching & 72.85 & 0.25 & & & 1051 \\
        Soft matching & 73.49 & 0.24 & & & \\
        \bottomrule
    \end{tabular}
\end{table}

\begin{table}[htbp]
    \small
    \centering
    \caption{Performance Metrics for the GPT3.5-turbo model on RE in materials-properties extraction. In this run, entities are shuffled to make it more challenging for the LLM. P: Precision, R: Recall, F1: harmonic average of P and R, Supp: Support, number of extracted entities.}
    \begin{tabular}{lccccc}
        \toprule
        \textbf{Run} & \textbf{Matching} & \textbf{P} & \textbf{R} & \textbf{F1} & \textbf{Supp} \\
        \midrule
        \multirow{2}{*}{Run1} & Strict matching & 72.66 & 66.49 & 69.44 & 1046 \\
        & Soft matching & 73.52 & 67.28 & 70.26 & 1046 \\
        \midrule
        \multirow{2}{*}{Run2} & Strict matching & 72.61 & 65.18 & 68.7 & 1026 \\
        & Soft matching & 74.07 & 66.49 & 70.08 & 1026 \\
        \midrule
        \multirow{2}{*}{Run3} & Strict matching & 73.16 & 66.05 & 69.43 & 1032 \\
        & Soft matching & 74.42 & 67.19 & 70.62 & 1032 \\
        \midrule
        \multicolumn{2}{l}{\textbf{Mean and Standard deviation of F1 score}} & & & & \\
        \midrule
        \textbf{Matching} & \textbf{Avg.} & $\sigma\textsuperscript{2}$ & & & \textbf{Avg. Supp}\\
        Strict matching & 72.85 & 0.25 & & & 1051 \\
        Soft matching   & 73.49 & 0.24 & & \\
        \bottomrule
    \end{tabular}
\end{table}

\begin{table}[htbp]
    \small
    \centering
    \caption{Performance Metrics for the GPT4 model on RE in materials-properties extraction. In this run, entities are not shuffled when added to the prompt. P: Precision, R: Recall, F1: harmonic average of P and R, Supp: Support, number of extracted entities.}
    \begin{tabular}{lccccc}
        \toprule
        \textbf{Run} & \textbf{Matching} & \textbf{P} & \textbf{R} & \textbf{F1} & \textbf{Supp} \\
        \midrule
        \multirow{2}{*}{Run1} & Strict matching & 82.41 & 74.19 & 78.08 & 1029 \\
        & Soft matching & 82.41 & 74.19 & 78.08 & 1029 \\
        \midrule
        \multirow{2}{*}{Run2} & Strict matching & 83.32 & 73.84 & 78.29 & 1013 \\
        & Soft matching & 83.51 & 74.02 & 78.48 & 1013 \\
        \midrule
        \multirow{2}{*}{Run3} & Strict matching & 83.5 & 73.93 & 78.42 & 1012 \\
        & Soft matching & 83.6 & 74.02 & 78.52 & 1012 \\
        \midrule
        \multicolumn{2}{l}{\textbf{Mean and Standard deviation of F1 score}} & & & & \\
        \midrule
        \textbf{Matching} & \textbf{Avg.} & $\sigma\textsuperscript{2}$ & & & \textbf{Avg. Supp}\\
        Strict matching & 78.26 & 0.17 & & & 1018 \\
        Soft matching & 78.36 & 0.24 & & \\
        \bottomrule
    \end{tabular}
\end{table}


\begin{table}[htbp]
    \small
    \centering
    \caption{Performance Metrics for the GPT4 model on RE in materials-properties extraction. In this run, entities are shuffled to make the task more challenging for the LLM. P: Precision, R: Recall, F1: harmonic average of P and R, Supp: Support, number of extracted entities.}
    \begin{tabular}{lccccc}
        \toprule
        \textbf{Run} & \textbf{Matching} & \textbf{P} & \textbf{R} & \textbf{F1} & \textbf{Supp} \\
        \midrule
        \multirow{2}{*}{Run1} & Strict matching & 81.19 & 72.88 & 76.81 & 1026 \\
        & Soft matching & 81.19 & 72.88 & 76.81 & 1026 \\
        \midrule
        \multirow{2}{*}{Run2} & Strict matching & 81.19 & 72.88 & 76.81 & 1026 \\
        & Soft matching & 81.19 & 72.88 & 76.81 & 1026 \\
        \midrule
        \multirow{2}{*}{Run3} & Strict matching & 82.77 & 74.37 & 78.34 & 1027 \\
        & Soft matching & 82.86 & 74.45 & 78.43 & 1027 \\
        \midrule
        \multicolumn{2}{l}{\textbf{Mean and Standard deviation of F1 score}} & & & & \\
        \midrule
        \textbf{Matching} & \textbf{Avg.} & $\sigma\textsuperscript{2}$ & & & \textbf{Avg. Supp}\\
        Strict matching & 77.46 & 0.78 & & & 1022 \\
        Soft matching & 77.52 & 0.80 & & \\
        \bottomrule
    \end{tabular}
\end{table}


\begin{table}[htbp]
    \small
    \centering
    \caption{Performance Metrics for the GPT4-turbo model on RE in materials-properties extraction. In this run, entities are not shuffled when they are added to the prompt. P: Precision, R: Recall, F1: harmonic average of P and R, Supp: Support, number of extracted entities.}
    \begin{tabular}{lccccc}
        \toprule
        \textbf{Run} & \textbf{Matching} & \textbf{P} & \textbf{R} & \textbf{F1} & \textbf{Supp}\\
        \midrule
        \multirow{2}{*}{Run1} & Strict matching & 86.48 & 71.65 & 78.37 & 947 \\
        & Soft matching & 87.54 & 72.53 & 79.33 & 947 \\
        \midrule
        \multirow{2}{*}{Run2} & Strict matching & 86.01 & 71.57 & 78.13 & 951 \\
        & Soft matching & 86.01 & 71.57 & 78.13 & 951 \\
        \midrule
        \multirow{2}{*}{Run3} & Strict matching & 86.67 & 71.13 & 78.14 & 938 \\
        & Soft matching & 87.53 & 71.83 & 78.9 & 938 \\
        \midrule
        \multicolumn{2}{l}{\textbf{Mean and Standard deviation of F-Score}} & & & & \\
        \midrule
        \textbf{Matching} & \textbf{Avg.} & $\sigma\textsuperscript{2}$ & & & \textbf{Avg. Supp}\\
        Strict matching & 78.21 & 0.13 & & & 945 \\
        Soft matching & 79.04 & 0.25 & & \\
        \bottomrule
    \end{tabular}
\end{table}


\begin{table}[htbp]
    \small
    \centering
    \caption{Performance Metrics for the GPT4-turbo model on RE in materials-properties extraction. In this run, entities are shuffled to make the task more challenging for the LLM. P: Precision, R: Recall, F1: harmonic average of P and R, Supp: Support, number of extracted entities.}
    \begin{tabular}{lccccc}
        \toprule
        \textbf{Run} & \textbf{Matching} & \textbf{P} & \textbf{R} & \textbf{F1} & \textbf{Supp} \\
        \midrule
        \multirow{2}{*}{Run1} & Strict matching & 86.59	70.08	77.47	925 \\
        & Soft matching & 87.46	70.78	78.24	925 \\
        \midrule
        \multirow{2}{*}{Run2} & Strict matching & 87.17	70.17	77.75	920\\
        & Soft matching & 87.93	70.78	78.43	920 \\
        \midrule
        \multirow{2}{*}{Run3} & Strict matching & 86.15	69.12	76.7	917\\
        & Soft matching & 87.02	69.82	77.48	917 \\
        \midrule
        \multicolumn{2}{l}{\textbf{Mean and Standard deviation of F-Score}} & & & & \\
        \midrule
        \textbf{Matching} & \textbf{Avg.} & $\sigma\textsuperscript{2}$ & & & \textbf{Avg. Supp}\\
        Strict matching & 77.30 & 0.54 & & & 920 \\
        Soft matching & 78.05 & 0.50 & & \\
        \bottomrule
    \end{tabular}
\end{table}

\clearpage
\subsubsection{Fine-tuning}

\begin{table}[htbp]
    \small
    \centering
    \caption{Performance Metrics for the GPT3.5-turbo model fine-tuned using the strategy "FT.base" on RE in materials-properties extraction. In this run, entities are not shuffled when included in the prompt. P: Precision, R: Recall, F1: harmonic average of P and R, Supp: Support, number of extracted entities.}
    \begin{tabular}{lccccc}
        \toprule
        \textbf{Run} & \textbf{Matching} & \textbf{P} & \textbf{R} & \textbf{F1} & \textbf{Supp} \\
        \midrule
        \multirow{2}{*}{Run1} & Strict matching & 91.06 & 80.00 & 85.17 & 123 \\
        & Soft matching & 93.50 & 82.14 & 87.45 & 123 \\
        \midrule
        \multirow{2}{*}{Run2} & Strict matching & 91.06 & 80.00 & 85.17 & 123 \\
        & Soft matching & 93.50 & 82.14 & 87.45 & 123 \\
        \midrule
        \multirow{2}{*}{Run3} & Strict matching & 91.06 & 80.00 & 85.17 & 123 \\
        & Soft matching & 93.50 & 82.14 & 87.45 & 123 \\
        \midrule
        \multicolumn{2}{l}{\textbf{Mean and Standard deviation of F1 score}} & & & & \\
        \midrule
        \textbf{Matching} & \textbf{Avg.} & $\sigma\textsuperscript{2}$ & & & \textbf{Avg. Supp}\\
        Strict matching & 85.17 & 0 &  & &  123 \\
        Soft matching & 87.45 & 0 & \\
        \bottomrule
    \end{tabular}
\end{table}

\begin{table}[htbp]
    \small
    \centering
    \caption{Performance Metrics for the GPT3.5-turbo model fine-tuned using the strategy "FT.base" on RE in materials-properties extraction. In this run, entities are shuffled to make it more challenging for the LLM. P: Precision, R: Recall, F1: harmonic average of P and R, Supp: Support, number of extracted entities.}
    \begin{tabular}{lccccc}
        \toprule
        \textbf{Run} & \textbf{Matching} & \textbf{P} & \textbf{R} & \textbf{F1} & \textbf{Supp} \\
        \midrule
        \multirow{2}{*}{Run1} & Strict matching & 92.68 & 81.43 & 86.69 & 123 \\
        & Soft matching & 95.12 & 83.57 & 88.97 & 123 \\
        \midrule
        \multirow{2}{*}{Run2} & Strict matching & 91.8 & 80 & 85.5 & 122 \\
        & Soft matching & 94.26 & 82.14 & 87.79 & 122 \\
        \midrule
        \multirow{2}{*}{Run3} & Strict matching & 91.06 & 80 & 85.17 & 123 \\
        & Soft matching & 93.5 & 82.14 & 87.45 & 123 \\
        \midrule
        \multicolumn{2}{l}{\textbf{Mean and Standard deviation of F1 score}} & & & & \\
        \midrule
        \textbf{Matching} & \textbf{Avg.} & $\sigma\textsuperscript{2}$ & & &  \textbf{Avg. Supp}\\
        Strict matching & 85.78 & 0.79 & & & 123 \\
        Soft matching & 88.07 & 0.79 & & \\
        \bottomrule
    \end{tabular}
\end{table}

\begin{table}[htbp]
    \small
    \centering
    \caption{Performance Metrics for the GPT3.5-turbo model fine-tuned using the strategy "FT.document\_order" on RE in materials-properties extraction. In this run, entities are not shuffled when included in the prompt. P: Precision, R: Recall, F1: harmonic average of P and R, Supp: Support, number of extracted entities.}
    \begin{tabular}{lccccc}
        \toprule
        \textbf{Run} & \textbf{Matching} & \textbf{P} & \textbf{R} & \textbf{F1} & \textbf{Supp} \\
        \midrule
        \multirow{2}{*}{Run1}   & Strict matching   & 88.98 & 80.71 & 84.64 & 127 \\
                                & Soft matching     & 91.34 & 82.86 & 86.89 & 127 \\
        \midrule
        \multirow{2}{*}{Run2}   & Strict matching   & 88.98 & 80.71 & 84.64 & 127 \\
                                & Soft matching     & 91.34 & 82.86 & 86.89 & 127 \\
        \midrule
        \multirow{2}{*}{Run3}   & Strict matching   & 88.98 & 80.71 & 84.64 & 127 \\
                                & Soft matching     & 91.34 & 82.86 & 86.89 & 127 \\
        \midrule
        \multicolumn{2}{l}{\textbf{Mean and Standard deviation of F1 score}} & & & & \\
        \midrule
        \textbf{Matching} & \textbf{Avg.} & $\sigma\textsuperscript{2}$ & & & \textbf{Avg. Supp}\\
        Strict matching & 84.64 & 0 &  & &  123 \\
        Soft matching   & 86.89 & 0 & \\
        \bottomrule
    \end{tabular}
\end{table}

\begin{table}[htbp]
    \small
    \centering
    \caption{Performance Metrics for the GPT3.5-turbo model fine-tuned using the strategy "FT.document\_order" on RE in materials-properties extraction. In this run, entities are shuffled to make it more challenging for the LLM. P: Precision, R: Recall, F1: harmonic average of P and R, Supp: Support, number of extracted entities.}
    \begin{tabular}{lccccc}
        \toprule
        \textbf{Run} & \textbf{Matching} & \textbf{P} & \textbf{R} & \textbf{F1} & \textbf{Supp} \\
        \midrule
        \multirow{2}{*}{Run1}   & Strict matching   & 47.73 & 45 & 46.32 & 132 \\
                                & Soft matching     & 48.48 & 45.71  & 47.06 & 132 \\
        \midrule
        \multirow{2}{*}{Run2}   & Strict matching   & 55.73 & 52.14  & 53.87 & 131 \\
                                & Soft matching     & 55.73 & 52.14  & 53.87 & 131 \\
        \midrule
        \multirow{2}{*}{Run3}   & Strict matching   & 57.72 & 50.71  & 53.99 & 123 \\
                                & Soft matching     & 57.72 & 50.71  & 53.99 & 123 \\
        \midrule
        \multicolumn{2}{l}{\textbf{Mean and Standard deviation of F1 score}} & & & & \\
        \midrule
        \textbf{Matching} & \textbf{Avg.} & $\sigma\textsuperscript{2}$ & & &  \textbf{Avg. Supp}\\
        Strict matching & 51.39 & 4.39 & & & 123 \\
        Soft matching & 51.64 & 3.96 & & \\
        \bottomrule
    \end{tabular}
\end{table}

\begin{table}[htbp]
    \small
    \centering
    \caption{Performance Metrics for the GPT3.5-turbo model fine-tuned using the strategy "FT.augmented" on RE in materials-properties extraction. In this run, entities are not shuffled when included in the prompt. P: Precision, R: Recall, F1: harmonic average of P and R, Supp: Support, number of extracted entities.}
    \begin{tabular}{lccccc}
        \toprule
        \textbf{Run} & \textbf{Matching} & \textbf{P} & \textbf{R} & \textbf{F1} & \textbf{Supp} \\
        \midrule
        \multirow{2}{*}{Run1}   & Strict matching   & 89.6 & 80 & 84.53 & 125 \\
                                & Soft matching     & 92 & 82.14 & 86.79 & 125 \\
        \midrule
        \multirow{2}{*}{Run2}   & Strict matching   & 89.6 & 80 & 84.53 & 125 \\
                                & Soft matching     & 92 & 82.14 & 86.79 & 125 \\
        \midrule
        \multirow{2}{*}{Run3}   & Strict matching   & 89.6 & 80 & 84.53 & 125 \\
                                & Soft matching     & 92 & 82.14 & 86.79 & 125 \\
        \midrule
        \multicolumn{2}{l}{\textbf{Mean and Standard deviation of F1 score}} & & & & \\
        \midrule
        \textbf{Matching} & \textbf{Avg.} & $\sigma\textsuperscript{2}$ & & & \textbf{Avg. Supp}\\
        Strict matching & 84.53 & 0 &  & &  125 \\
        Soft matching   & 86.79 & 0 & \\
        \bottomrule
    \end{tabular}
\end{table}

\begin{table}[htbp]
    \small
    \centering
    \caption{Performance Metrics for the GPT3.5-turbo model fine-tuned using the strategy "FT.augmented" on RE in materials-properties extraction. In this run, entities are shuffled to make it more challenging for the LLM. P: Precision, R: Recall, F1: harmonic average of P and R, Supp: Support, number of extracted entities.}
    \begin{tabular}{lccccc}
        \toprule
        \textbf{Run} & \textbf{Matching} & \textbf{P} & \textbf{R} & \textbf{F1} & \textbf{Supp} \\
        \midrule
        \multirow{2}{*}{Run1}   & Strict matching   & 91.13 & 80.71 & 85.61 & 124 \\
                                & Soft matching     & 91.94 & 81.43 & 86.36 & 124 \\
        \midrule
        \multirow{2}{*}{Run2}   & Strict matching   & 87.9 & 77.86 & 82.58 & 124 \\
                                & Soft matching     & 88.71 & 78.57 & 83.33 & 124 \\
        \midrule
        \multirow{2}{*}{Run3}   & Strict matching   & 89.52 & 79.29 & 84.09 & 124 \\
                                & Soft matching     & 91.94 & 81.43 & 86.36 & 124 \\
        \midrule
        \multicolumn{2}{l}{\textbf{Mean and Standard deviation of F1 score}} & & & & \\
        \midrule
        \textbf{Matching} & \textbf{Avg.} & $\sigma\textsuperscript{2}$ & & &  \textbf{Avg. Supp}\\
        Strict matching & 84.09 & 1.51 & & & 124 \\
        Soft matching & 85.35 & 1.74 & & \\
        \bottomrule
    \end{tabular}
\end{table}

\end{document}